\documentclass[pdflatex,sn-nature]{sn-jnl}

\makeatletter
\@twosidefalse
\@mparswitchfalse
\makeatother
\usepackage{graphicx}
\usepackage{multirow}
\usepackage{amsmath,amssymb,amsfonts}
\usepackage{amsthm}
\usepackage{mathrsfs}
\usepackage[title]{appendix}
\usepackage{xcolor}
\usepackage{textcomp}
\usepackage{manyfoot}
\usepackage{booktabs}
\usepackage{wrapfig}

\newcommand{\name}{\texttt{TextRadar}}
\newcommand{\header}[1]{\vskip 0.1cm \noindent{\bf #1}}

\newcommand{\figref}[1]{Fig.~\ref{#1}}
\newcommand{\tabref}[1]{Table~\ref{#1}}
\newcommand{\secref}[1]{Section~\ref{#1}}

\begin{document}
\title[3D Occupancy-Language Fields from mmWave Radar]{\protect\shortstack[c]{3D Occupancy-Language Fields from mmWave Radar}}

\author*{Haowen Lai\textsuperscript{*}}\email{hwlai@seas.upenn.edu}
\author{Zhiwei Zheng}
\author*{Mingmin Zhao\textsuperscript{*}}\email{mingminz@seas.upenn.edu}
\affil{\orgname{University of Pennsylvania}, \orgaddress{\city{Philadelphia}, \state{PA}, \postcode{19104}, \country{USA}}}

\abstract{
Radar systems can recover scene geometry and recognize predefined categories, but these outputs do not provide a persistent spatial representation for open-ended language queries.
Such queries require linking observations from different viewpoints to the same physical locations.
We present \name{}, a framework that constructs a scene-level 3D occupancy-language field from panoramic radar observations.
The field combines occupancy with scale-dependent language features at surface locations, allowing the reconstructed scene to be queried with new descriptions.
A single-frame radar distillation model predicts aligned ranges and language features from each radar observation.
Using radar-estimated poses, we register these predictions to reconstruct the field from radar alone in a new scene.
Experiments on approximately 20k synchronized observations across seven buildings demonstrate scene-level natural-language querying and prompt-free semantic visualization.
Analysis of camera-to-radar concept transferability shows stronger agreement for structures, materials, and human poses than for colors and printed text.
}

\maketitle

\section{Introduction}
Recent years have seen rapid progress in mmWave sensing, transforming radars from coarse detectors into high-resolution imaging systems~\cite{panoradar,holoradar,cartoradar}.
Much of this progress has focused on resolving surfaces and recovering scene geometry~\cite{radatron,see_through_smoke,HawkEye}.
Geometry alone, however, cannot tell a robot what a sensed region contains or how it might be used.
This information matters for tasks such as region search~\cite{Scanrefer,Llm_grounder} and semantic navigation~\cite{asante2024toward,huang2024embodied}, where a robot may be asked to find ``a place to sit''.
Such a request could refer to a chair, a bench, or another suitable surface rather than one predefined object label.

RF sensing provides some of this semantic information by recognizing scene categories, materials, and human activities~\cite{panoradar,see_through_smoke,surfradar,Soli,waveverse}.
These systems generally choose from predefined labels, which limits the descriptions a person can use to search a scene.
Recent radar-language models expand this range by generating descriptions of human motion and scene content~\cite{radarllm,HoloLLM,hamilton2026weather}.
Their language, however, is usually tied to an input frame, scene, or motion sequence rather than to physical locations in a shared 3D map.
For a moving robot, that spatial link matters: a description of a chair in one view does not tell it where to find the chair later or whether another view shows the same one.
Answering such questions requires associating observations from different viewpoints with persistent 3D locations.
Vision-based language fields show how this kind of representation supports spatial queries~\cite{lerf,Langsplat}.
Can radar provide a comparable representation that links language to reconstructed surfaces across a scene?

\begin{figure*}[t]
\centering
\includegraphics[width=\linewidth]{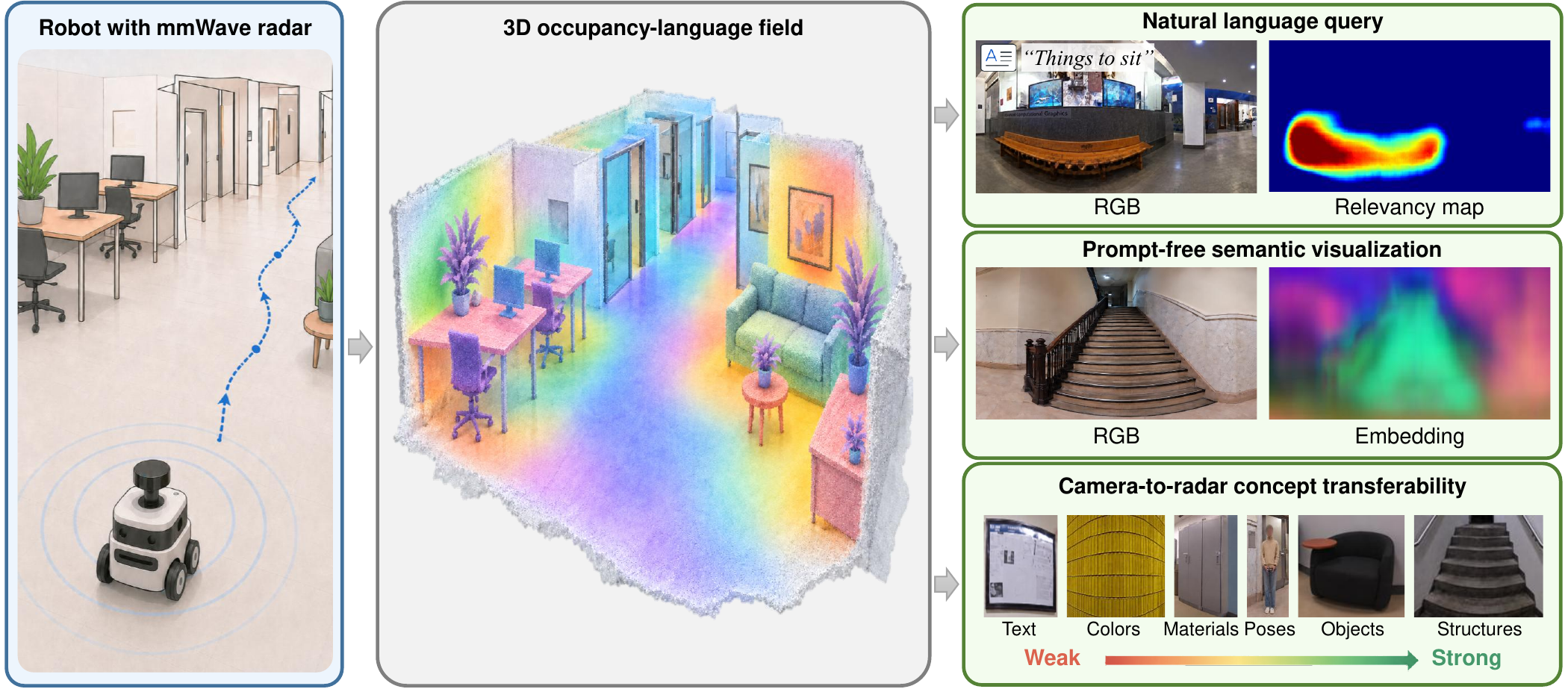}
\caption{\textbf{TextRadar constructs a scene-level 3D occupancy-language field from radar observations.}
A robot collects panoramic mmWave measurements along a trajectory (left).
Surface geometry and language features predicted from these measurements are integrated in shared 3D coordinates (center).
The right panels illustrate natural-language querying, prompt-free semantic visualization, and camera-to-radar concept transferability.
The query ``Things to sit'' highlights a bench, while the feature visualization reveals semantic organization without specifying a prompt.
Comparisons of radar-predicted and camera-derived features show stronger agreement for structures, objects, materials, and human poses than for colors and printed text.}
\label{fig:teaser_fig}
\end{figure*}

To address this question, we present \name{}, a framework for constructing a scene-level \emph{3D occupancy-language field} from panoramic radar observations.
The field represents scene occupancy and associates surface locations with language features at multiple context scales.
As illustrated in \figref{fig:teaser_fig}, a robot gathers panoramic radar measurements along a trajectory.
For each observation, a learned model predicts distances to surfaces and spatially aligned language features.
Camera and LiDAR measurements provide training targets for the model.
In a new scene, the trained model predicts both quantities from radar alone.
Using radar-estimated poses, \name{} places these predictions in shared 3D coordinates and combines repeated observations at the same locations.
The resulting field retains surfaces and their language features after the robot leaves a region, allowing users to search with new descriptions without rebuilding the map.
For example, ``things to sit'' retrieves relevant seating, illustrated by the highlighted bench in \figref{fig:teaser_fig}.
The same features also support prompt-free semantic visualization, revealing how different scene regions are organized.
Finally, we compare radar-predicted features with camera-derived references to assess camera-to-radar concept transferability and characterize the language information available to build the field.

Realizing this framework requires three components: spatial language distillation, panoramic supervision, and reconstruction from radar predictions.

\header{Distilling spatial language features from vision.}
To support natural-language querying after reconstruction, a scene-level language field needs features tied to physical locations rather than a single label or caption for an observation.
Pretrained image-text models provide language features from camera images, but each feature describes an image region rather than a radar return.
We extract overlapping crops at multiple scales so the features describe both local objects and broader scene context.
To associate these camera features with radar-observed surfaces, we project LiDAR points into the image using calibrated sensor geometry.
At each projected point and crop scale, we bilinearly interpolate the embeddings at the four neighboring crop centers.
LiDAR depth also lets us express the physical area covered by each crop, so its context scale can be compared across distances.
A single-frame radar distillation model learns to predict these language features alongside LiDAR-supervised ranges on the same panoramic grid.
Its aligned outputs provide the surface locations and language observations needed to build the 3D field, without manual RF-language annotations.

\header{Extending supervision to the radar panorama.}
The camera provides perspective views of a limited sector, whereas each radar observation spans the full surrounding panorama.
Training only on that sector leaves other directions without language targets and biases learning toward the camera's viewing direction.
Without supervision elsewhere, the model may rely on direction-specific shortcuts that do not generalize across the panorama.
This can produce weak or spurious language features outside the supervised sector, leading to artifacts and uneven semantic coverage in the reconstructed scene.
Rotating training samples changes their coordinates but cannot supply the missing targets or remove this bias.
To extend coverage, we use robot motion and LiDAR geometry to accumulate camera-derived features in a 3D reference map.
For each panoramic surface point, we retrieve nearby features observed at compatible physical context scales.
This brings camera observations from different viewpoints into a common training target.
The resulting panoramic targets supervise the distillation model alongside LiDAR ranges, extending language learning beyond the camera's instantaneous view.

\header{Reconstructing the field from radar predictions.}
Once trained, the distillation model predicts ranges and language features from each radar observation in a new scene.
We use radar-estimated poses to register these predictions from different viewpoints in shared 3D coordinates.
The ranges constrain occupancy, while the language features supervise the language field at the corresponding surface points.
At the same point, however, language features can describe different amounts of context.
For example, one feature may describe a chair and another its surrounding seating area, so merging them into one embedding would lose that distinction.
Following LERF~\cite{lerf}, we condition the language field on spatial context scale.
We use predicted ranges to express feature scales in physical units and interpolate between neighboring scales when fitting the field.
Repeated observations thus constrain the same occupancy value and, at a given location and scale, the same language embedding.
This shared field retains the reconstructed surfaces and their language features for later queries, including regions outside the robot's current view.

We implement \name{} on a mobile robot equipped with a rotating mmWave radar, a front-facing camera, and a 3D LiDAR.
We collect approximately 20k synchronized observations along 21 trajectories across seven buildings.
To test generalization, we hold out each building in turn and train the distillation model on the other six.
Across the held-out buildings, the model achieves a mean range error of 16.38~cm and a mean radar-camera feature cosine similarity of 91.92\%.
These measurements describe the accuracy of its surface distances and the agreement of its language features with camera-derived references.
Using these predictions and radar-estimated poses, we reconstruct scene-level fields that support natural-language querying of extended structures and individual objects.
The same fields support prompt-free semantic visualization of the learned features.
We further evaluate camera-to-radar concept transferability with 42 descriptive prompts spanning objects and structures, materials, human poses, colors, and printed text.
For objects and structures, the mean agreement with the camera's preference between prompt pairs is 82.5\%, compared with a chance baseline of 50\%.
Together, the category-wise results show stronger agreement for structures, materials, and human poses than for colors and printed text, indicating which semantic cues are more reliably supplied to the field.

In summary, our contributions are:
\begin{itemize}
\item A scene-level 3D occupancy-language field that combines radar-derived surface reconstruction with multiscale language features for persistent spatial querying.
\item A single-frame radar distillation model and geometry-aware panoramic supervision that supply aligned range and language predictions for field construction.
\item Natural-language querying of structures and objects, together with prompt-free semantic visualization of the learned language features.
\item Evaluation on approximately 20k synchronized observations across seven buildings, showing stronger camera-to-radar concept agreement for structures, materials, and human poses than for colors and printed text.
\end{itemize}

\subsection{Related work}
\label{sec:related_work}
Image-text models such as CLIP~\cite{clip} provide a shared embedding space for visual content and natural language.
Building on this foundation, Distilled Feature Fields~\cite{dff} and LERF~\cite{lerf} embed pretrained image or language-aligned features into 3D fields for scene decomposition and open-vocabulary queries, while LangSplat~\cite{Langsplat} and Language Embedded 3D Gaussians~\cite{shi2024language} attach such features to Gaussian representations.
ConceptFusion~\cite{conceptfusion} and OpenScene~\cite{openscene} fuse multiview features into explicit 3D maps or point-cloud representations.
Visual Language Maps~\cite{huang23vlmaps} applies spatially grounded visual-language features to robot navigation.
Together, these methods establish language-aligned 3D scene understanding beyond predefined categories.
\name{} extends this capability to radar: camera and LiDAR data supervise the distillation model during training, while a new scene's occupancy and language fields are reconstructed from radar predictions and radar-estimated poses.

Producing these observations requires radar to recover both spatial detail and semantic cues.
Synthetic-aperture imaging exploits sensor motion to resolve surfaces and indoor layouts~\cite{panoradar,holoradar,dodds_mmnorm,UCSD_3D_pc,wei2021nonline,HawkEye,yanik2020development,avtwin,AIM_liliqiu}, while SLAM and 3D mapping aggregate measurements across viewpoints to reconstruct larger-scale environments~\cite{yin2023automerge,cartoradar,lai2022adafusion,liang2022global,lai2021LiDAR}.
Besides these systems, learned reconstruction methods recover dense ranges, occupancy, human geometry, and wave propagation from sensor measurements~\cite{cartoradar,radatron,hydra_Swarun_Ashutosh,park2025diffusion_liliqiu,wang2025high_Liliqiu,GRT_Anthony,radarocc_Chris,m4esh,avr,versa}.
In particular, CartoRadar formulates scene reconstruction through learned occupancy prediction~\cite{cartoradar}.
Beyond geometry, wireless sensing systems recognize scene categories~\cite{panoradar,see_through_smoke,zhang2023spatial,Hao_2024_CVPR}, materials~\cite{surfradar,shang2022liqray,xie2019tagtag,jiao2024exploring}, activities~\cite{sigg2013leveraging,Soli,crowd_Ashutosh_Yasamin,yang2025leakyfeeder}, and vital signs~\cite{hu2025tracking,yang2026motion,yang2022localized,chen2024exploring,wu2025heart}.
However, neither geometric mapping and reconstruction nor task-specific recognition provides a language field that can be queried at arbitrary locations in a 3D scene.
\name{} connects these capabilities by integrating radar-predicted geometry and language embeddings within shared 3D coordinates.

Recent radar-language systems provide another part of this foundation.
Building on large language models~\cite{achiam2023gpt,sushil2024comparative}, they use language models for supervised detection~\cite{marine_llm_target_detection}, reason over RF features to identify materials~\cite{llmaterial,RF_ray}, or describe human motion and activity~\cite{radarllm,HoloLLM}.
More recently, vision-aligned 4D radar features have also been used to generate structured scene captions~\cite{hamilton2026weather}.
These systems connect radar to language through detections, property estimates, and descriptions.
\name{} instead distills spatially aligned multiscale language features and integrates them with occupancy in a persistent 3D field, allowing the same reconstructed scene to support different spatial queries.

\section{Results}
\label{sec:results}
We first examine the scene-level occupancy-language field through 3D reconstruction and natural-language querying, then inspect the features it integrates through prompt-free visualization and camera-to-radar concept transferability.
We subsequently evaluate the single-frame radar distillation model that supplies the field's observations.
Experiments use approximately 20k synchronized observations from 21 trajectories in seven buildings, with the hardware, dataset, and training protocol detailed at the end of this section.

\begin{figure*}[t]
\centering
\includegraphics[width=0.85\textwidth]{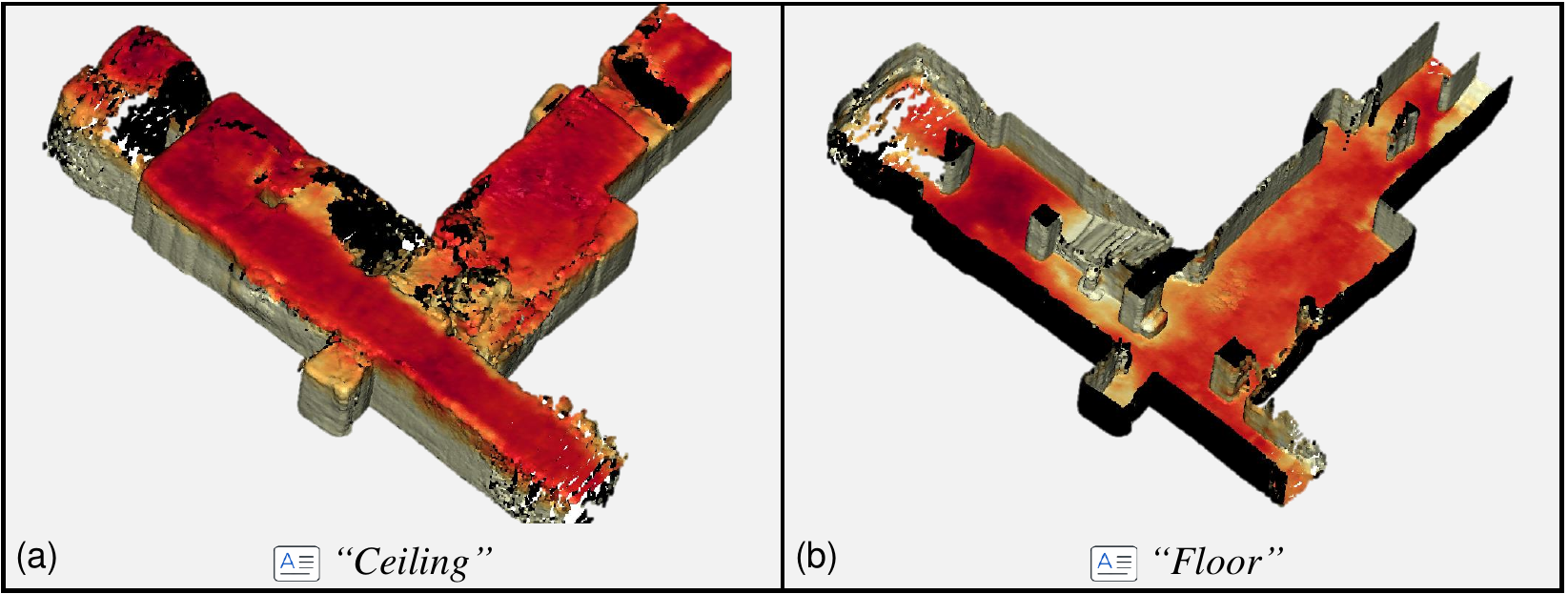}
\caption{\textbf{Scene-level natural-language querying of large building structures.}
The reconstructed geometry is colored according to the cosine similarity between the scene-level feature at each surface point and the CLIP text embedding of the query.
Warmer colors indicate higher semantic relevance.
The queries (a) ``Ceiling'' and (b) ``Floor'' activate complementary upper and lower surfaces across multiple corridors and room regions.
The spatially coherent responses over distant parts of the environment show how the persistent occupancy-language field associates related structures across a common 3D coordinate system.}
\label{fig:scene_large_structures}
\end{figure*}

\subsection{Scene-level geometry and natural-language querying}
\label{sec:scene_results}
\name{} reconstructs an indoor scene as a 3D occupancy-language field that can be queried after the robot has traversed the environment.
For these reconstructions, we use radar-predicted ranges, radar-predicted language features, and radar-estimated poses.
The field combines these observations in a common coordinate system, associating reconstructed surfaces with scale-dependent embeddings that can be compared with new text queries.
Once the field has been constructed, changing the query changes the regions retrieved from the scene without retraining either the distillation model or the field.

\begin{figure*}[t]
\centering
\includegraphics[width=\textwidth]{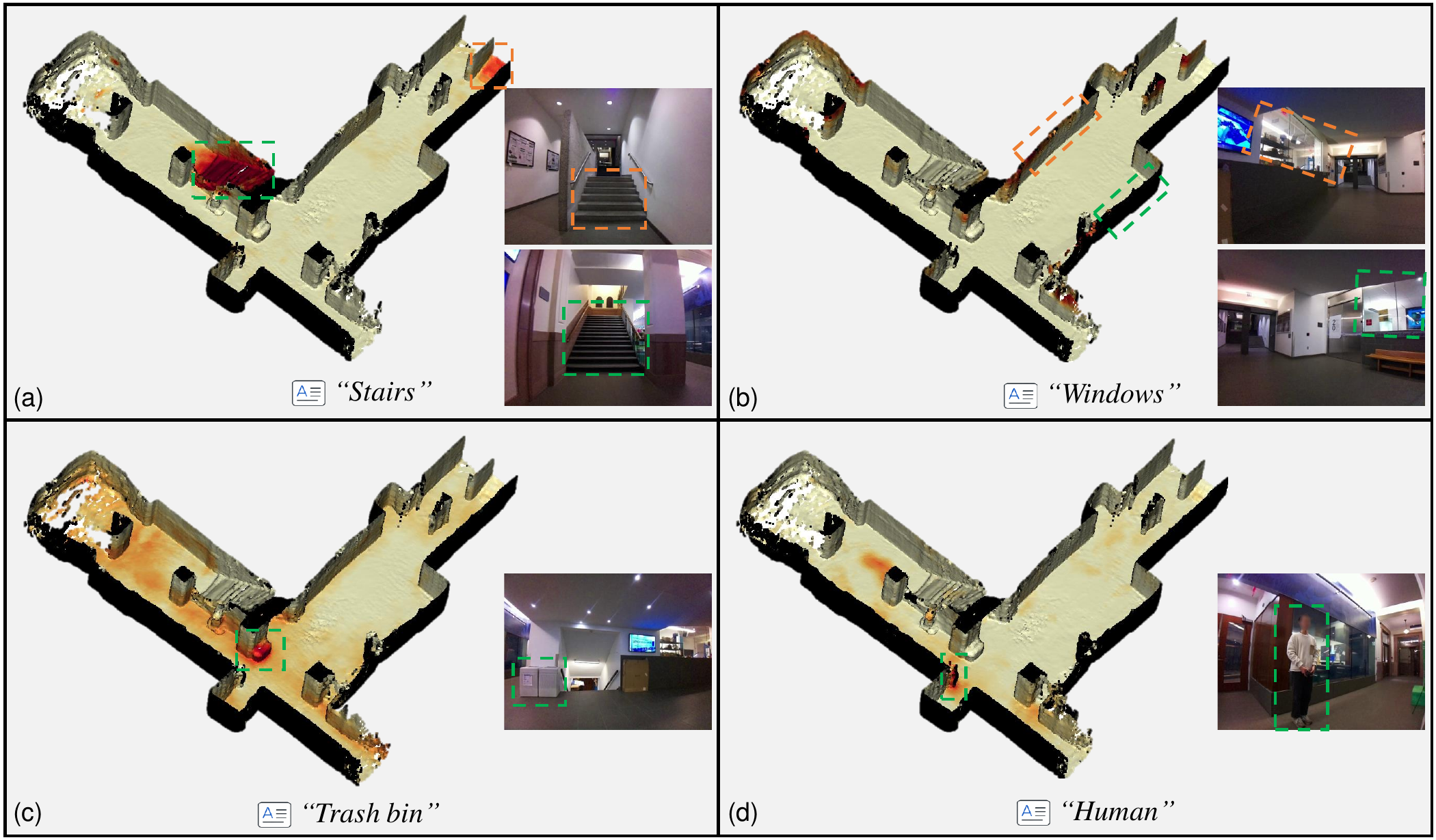}
\caption{\textbf{Scene-level natural-language querying of small structures and objects.}
The field localizes architectural structures, including (a) ``Stairs'' and (b) ``Windows,'' as well as compact objects, including (c) ``Trash bin'' and (d) ``Human.''
The RGB images provide representative observations of the queried regions, with dashed boxes indicating the corresponding targets.
For categories with multiple instances, such as stairs and windows, the system simultaneously highlights geometrically separated regions throughout the scene.}
\label{fig:scene_small_structures_objects}
\end{figure*}

For example, in \figref{fig:scene_large_structures}, ``Ceiling'' and ``Floor'' retrieve complementary upper and lower surfaces across the reconstructed corridors and adjoining regions.
The responses extend over areas collected from multiple sensor positions, making the queried structure accessible beyond any single radar panorama.
At a more local scale like \figref{fig:scene_small_structures_objects}, ``Stairs'' and ``Windows'' highlight multiple separated instances within the same scene.
The same field also produces responses to ``Trash bin'' and ``Human'' near the indicated objects, although some activation extends to other surfaces.
Together, these examples show that one field supports both queries for extended building structure and retrieval of small regions within that structure.

The common 3D coordinates are central to this behavior.
Specifically, each predicted range constrains the occupancy field along a radar ray, and its associated language features constrain the language field at the corresponding surface.
When a surface is revisited, observations constrain the same 3D field rather than remaining separate frame-indexed outputs.
The language field also takes context scale as input, allowing a location to represent both local structure and its broader surroundings.
For a fixed location and scale, the field returns the same embedding regardless of the viewpoint used to inspect it.
The query maps color the reconstructed geometry by cosine similarity to the CLIP text embedding; details of field construction and querying appear in \secref{sec:lang_field}.

\begin{figure*}[htbp]
\centering
\includegraphics[width=0.95\linewidth]{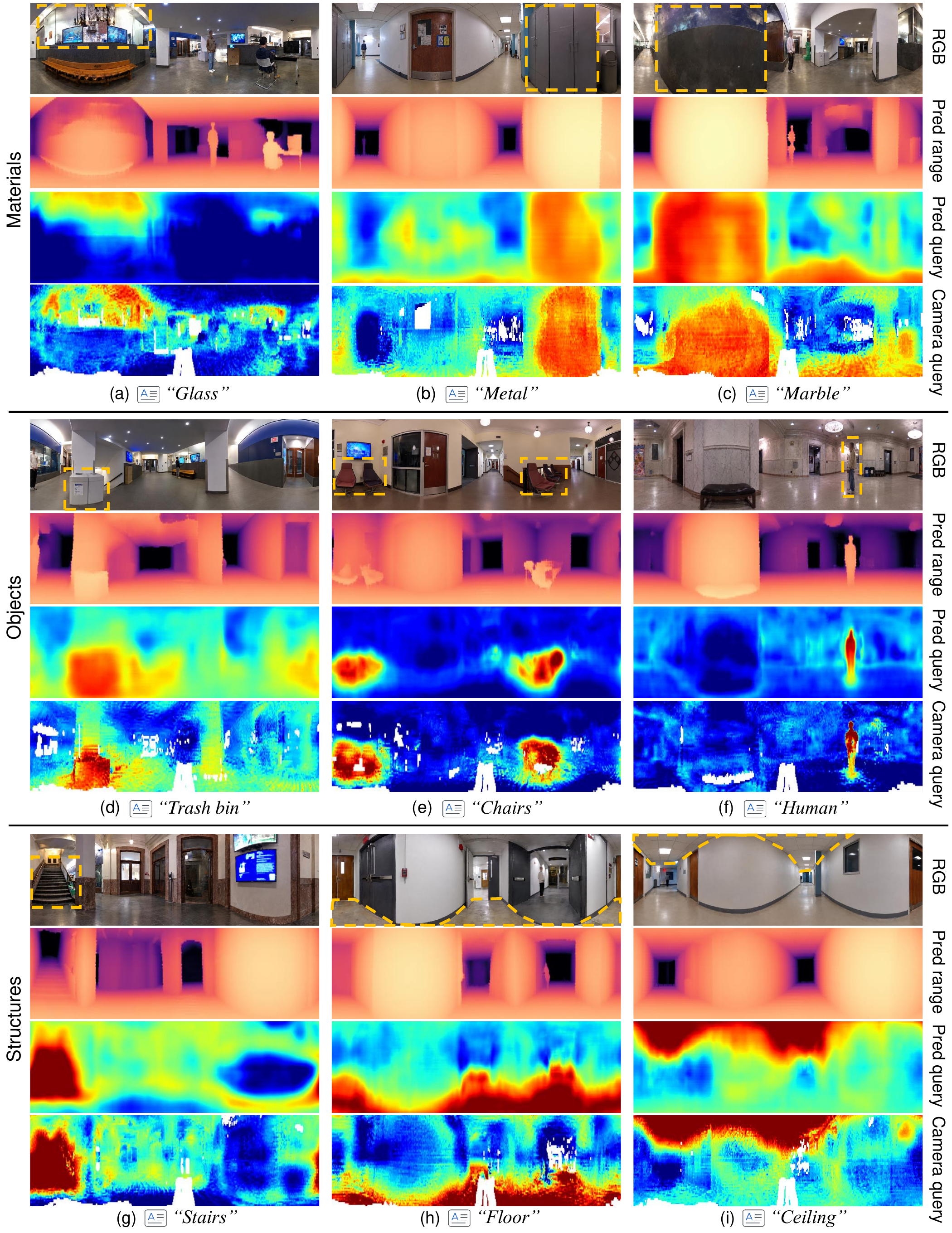}
\caption{\textbf{Geometry and natural-language queries.}
Each example shows, from top to bottom, the rendered panoramic RGB reference, the range image predicted from the radar measurement, the query-relevancy map obtained from the radar-predicted CLIP-aligned features, and the corresponding map obtained from the camera-derived reference features.
Warmer colors indicate higher cosine similarity to the text query.
The queries cover materials, including (a) ``Glass,'' (b) ``Metal,'' and (c) ``Marble''; objects, including (d) ``Trash bin,'' (e) ``Chairs,'' and (f) ``Human''; and structures, including (g) ``Stairs,'' (h) ``Floor,'' and (i) ``Ceiling.''
The radar-predicted maps localize both compact objects and extended structural regions, remain spatially consistent with the recovered geometry, and broadly agree with the camera-derived reference maps.}
\label{fig:single_frame_query}
\end{figure*}

To understand the evidence underlying these scene-level responses, we next inspect the frame-level predictions used to construct the field.
\figref{fig:single_frame_query} shows the predicted range image and language responses from individual radar panoramas, alongside camera-derived reference responses on the same grid.
The queries ``Glass,'' ``Metal,'' and ``Marble'' highlight the corresponding window, cabinet, and wall regions.
Similarly, ``Trash bin,'' ``Chairs,'' and ``Human'' produce localized responses, while ``Stairs,'' ``Floor,'' and ``Ceiling'' address larger structural regions.
The radar maps are smoother than the camera-derived references, but their dominant responses generally follow the locations and boundaries of the recovered surfaces.
These aligned ranges and features provide the observations for the scene field.

\begin{figure*}[t]
\centering
\includegraphics[width=0.95\linewidth]{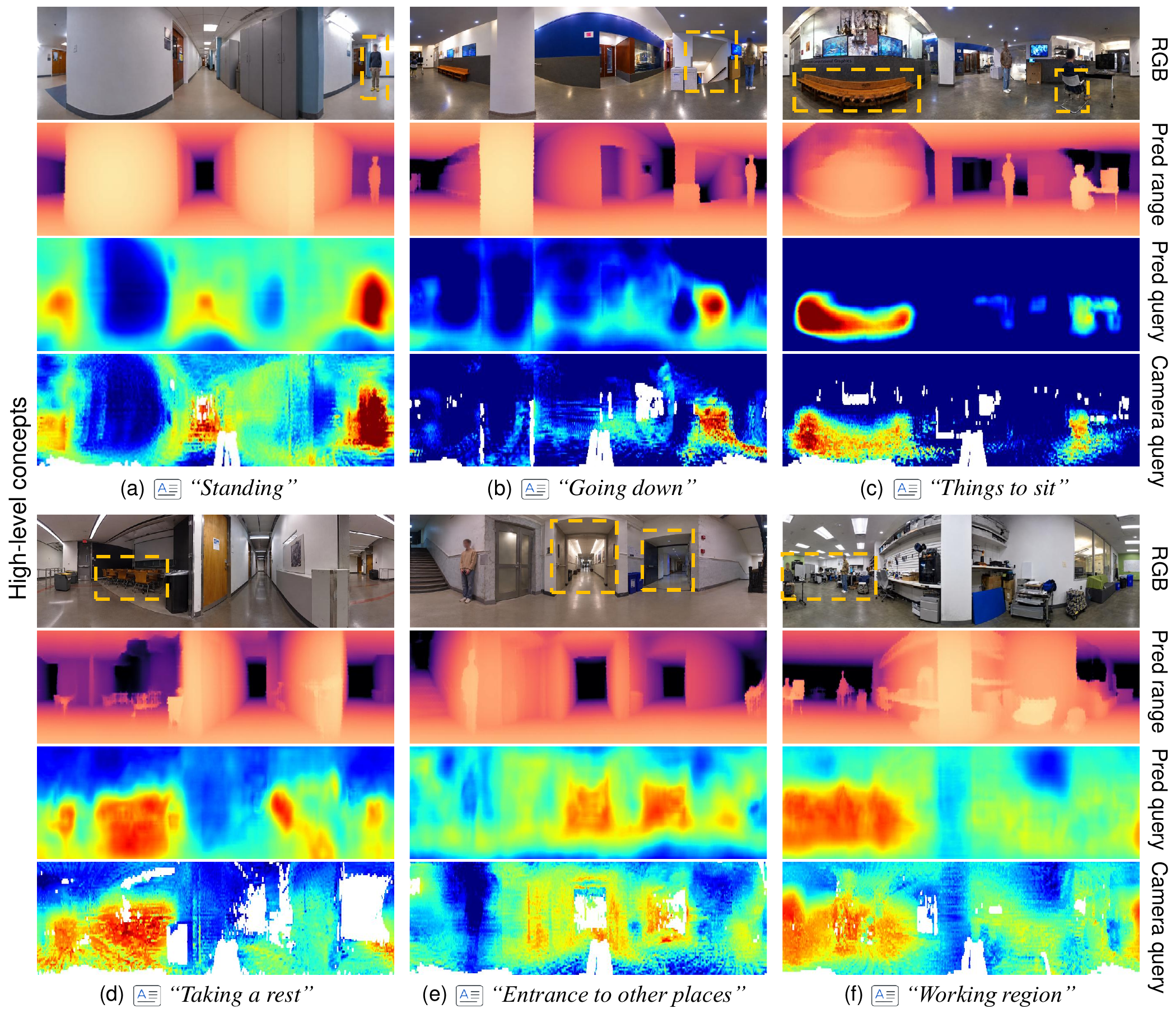}
\caption{\textbf{Natural-language querying of high-level concepts.}
The prompts describe human configuration, affordances, and functional regions: (a) ``Standing'' localizes a standing person, (b) ``Going down'' highlights a descending staircase, (c) ``Things to sit'' retrieves a bench and other seating, (d) ``Taking a rest'' identifies a shared seating area, (e) ``Entrance to other places'' highlights doorways and passages, and (f) ``Working region'' identifies the occupied workspace in a laboratory office.
These examples show responses to contextual prompts beyond direct object and material names.}
\label{fig:single_frame_query_high_level}
\end{figure*}

Beyond object and material names, the language features also support high-level contextual descriptions, as illustrated in \figref{fig:single_frame_query_high_level}.
For instance, ``Standing'' responds near an upright person, ``Going down'' highlights a descending staircase, and ``Things to sit'' retrieves seating.
At a broader contextual scale, ``Taking a rest,'' ``Entrance to other places,'' and ``Working region'' emphasize functional regions that span objects and their surroundings.
Together, these examples demonstrate responses to descriptive prompts beyond direct category names.
Such responses should be interpreted in their spatial context: ``Going down,'' for example, identifies a relevant structure and does not by itself establish recognition of an ongoing action.
The same multiscale feature space supports these queries at the observation level and supplies the language evidence used by the 3D field.

\subsection{Prompt-free semantic visualization}
\label{sec:embedding_visualization}
Natural language querying reveals the response to a selected prompt, but it does not show how the full feature space organizes a scene before a concept is specified.
We therefore inspect the organization of the language features that populate the occupancy-language field.
The comparisons below visualize the single-frame predictions used as field observations, where radar and camera-derived features can be compared at corresponding pixels.
For each comparison, we concatenate the radar-predicted features and the corresponding camera-derived reference features over all valid pixels, and fit a shared principal component analysis (PCA) projection from the original 768-dimensional feature space to three dimensions.
We then map the three projected coordinates to the red, green, and blue image channels.
Because the predicted and reference features share the same PCA basis, similar colors indicate proximity within their common three-dimensional projection.
In the comparisons shown here, however, a separate PCA basis is fitted to each prediction--reference pair, so the component orientations and corresponding color axes are arbitrary.
Colors can therefore be compared within a pair, but not across different examples.
We call this semantic visualization because it exposes structure in a language-aligned feature space.
The projection requires no text prompts, semantic labels, or predefined classes.
For scene-level visualization, the same operation can be applied to features sampled on the reconstructed 3D surfaces, with one PCA basis kept fixed for the entire scene and all its views.

\begin{figure*}[t]
\centering
\includegraphics[width=\textwidth]{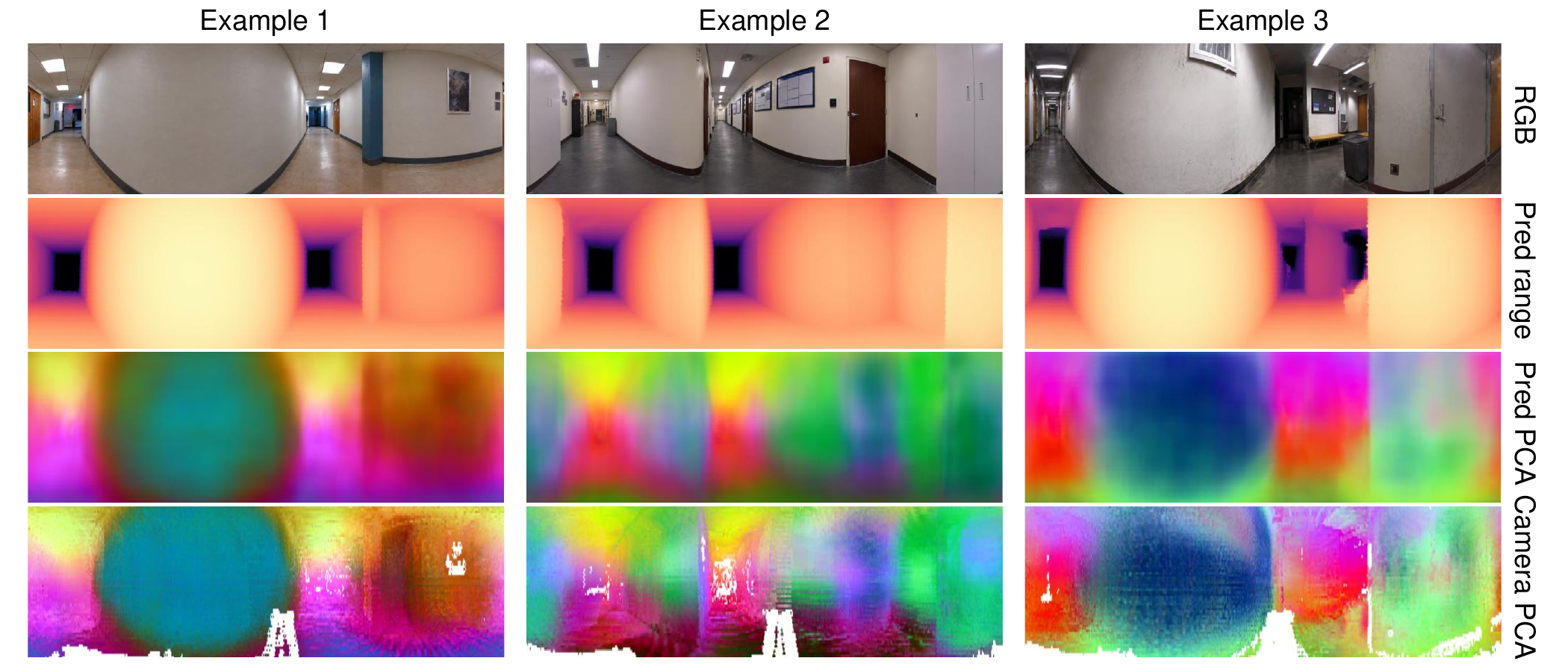}
\caption{\textbf{Prompt-free semantic visualization of general indoor structures.}
For each example, the rows from top to bottom present the rendered panoramic RGB image, the predicted range image, the three-component PCA projection of the radar-predicted CLIP features, and the corresponding camera-derived reference projection.
The predicted and reference features are projected using a shared PCA basis, and the three principal components are mapped to RGB channels.
Similar colors therefore indicate nearby positions in the projected semantic feature space, rather than predefined semantic classes.
Large structural regions, including ceilings, floors, walls, corridor openings, and wall-mounted structures, form spatially coherent groups in the radar-predicted visualization and broadly agree with the camera-derived reference.}
\label{fig:cluster_general_structure}
\end{figure*}

\begin{table*}[t]
\centering
\caption{\textbf{Category-wise radar-camera concept agreement.}
Camera-derived CLIP relevancy is the reference signal, and every property is evaluated within the entity that carries it.
Values are means over the seven held-out buildings and the complete prompt bank.}
\label{tab:concept_capture}
\resizebox{\textwidth}{!}{%
\begin{tabular}{llcccc}
\toprule
 & & & Pairwise & Region & Instance-level \\
Category & Evaluation entity & \# prompts & discrimination $\uparrow$ & AUROC $\uparrow$ & correlation $\uparrow$ \\
 & & & (chance 0.50) & (chance 0.50) & (null 0.00) \\
\midrule
Objects / structures & whole frame & 16 & 0.825 & 0.859 & 0.885 \\
Materials & one object class & 6 & 0.856 & 0.694 & 0.542 \\
Human actions & person & 6 & 0.755 & 0.704 & 0.552 \\
\midrule
Colors & one object class & 9 & 0.669 & 0.592 & 0.273 \\
Text & display & 5 & 0.579 & 0.604 & 0.277 \\
\bottomrule
\end{tabular}
}
\end{table*}

\begin{figure*}[t]
\centering
\includegraphics[width=\textwidth]{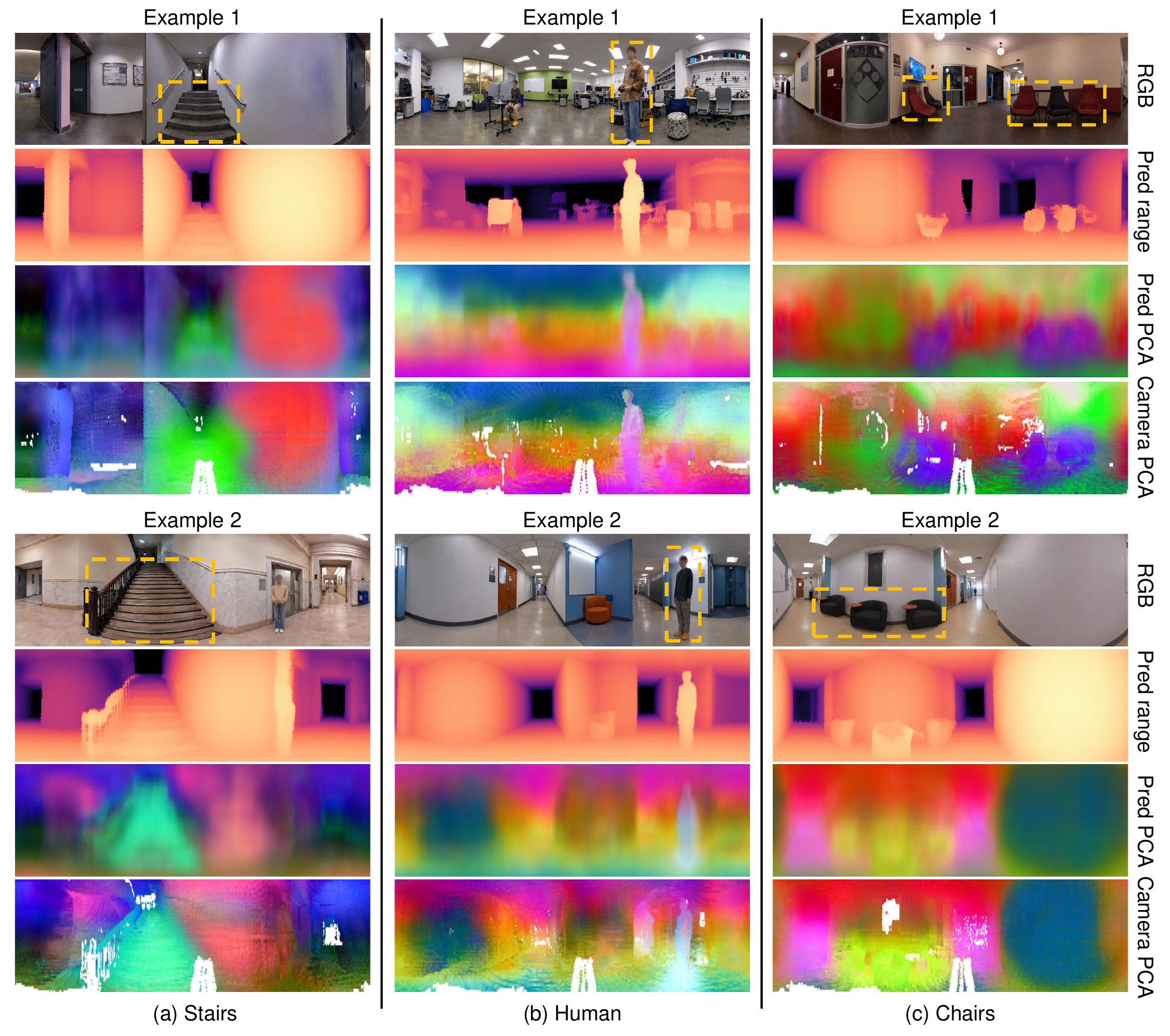}
\caption{\textbf{Prompt-free semantic visualization of structures and objects.}
The examples examine (a) stairs, (b) humans, and (c) chairs, with two scenes shown for each concept.
The dashed boxes in the rendered RGB image indicate the structures or objects of interest.
A shared PCA projection is used for each prediction-reference comparison, allowing their color organization to be compared directly.
Stairs, human silhouettes, and individual chairs emerge as coherent regions whose boundaries follow the corresponding geometry, despite some differences in their sizes, shapes, and spatial extents.}
\label{fig:cluster_objects}
\end{figure*}

\begin{figure*}[t]
\centering
\includegraphics[width=\textwidth]{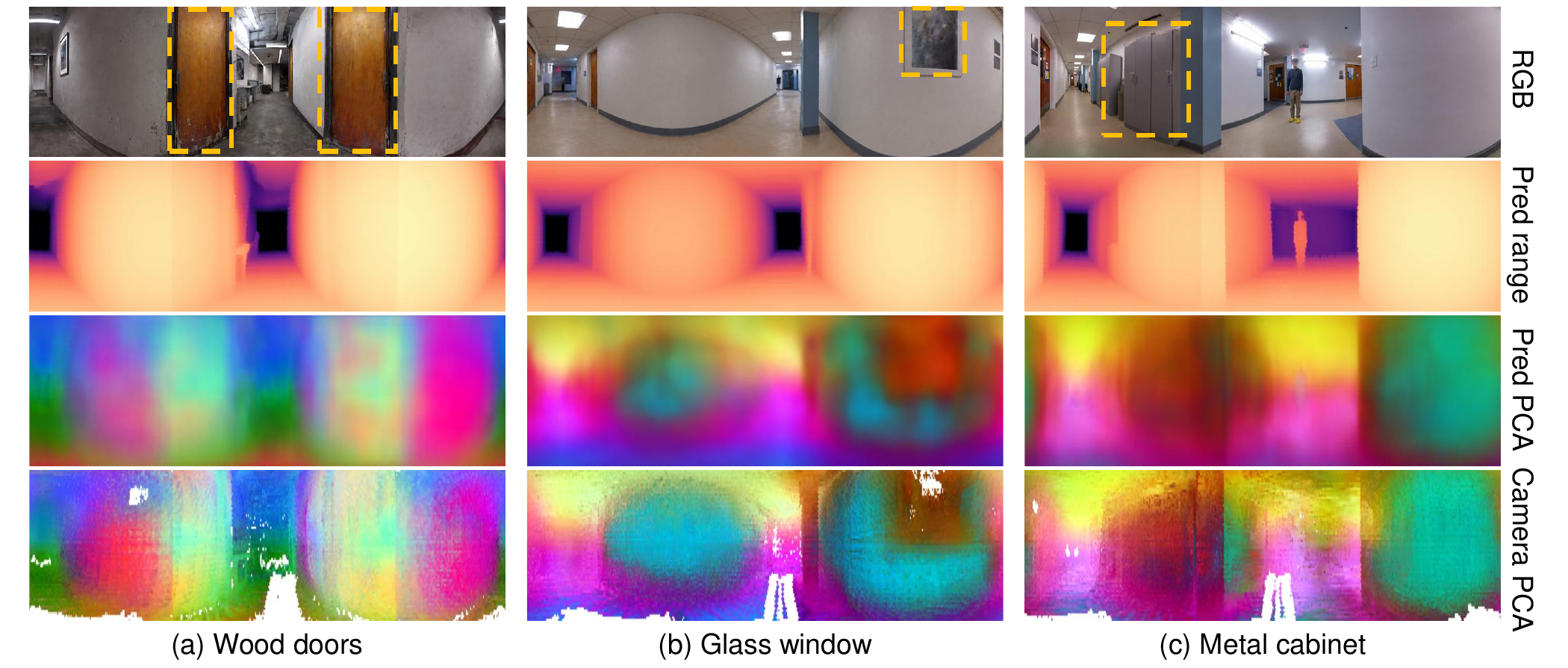}
\caption{\textbf{Prompt-free semantic visualization of material-related regions.}
The examples show (a) wood doors, (b) a glass window, and (c) a metal cabinet.
The highlighted material regions occupy distinct portions of the projected feature space and therefore appear with colors different from their surrounding walls and structures.
The agreement between the radar-predicted and camera-derived PCA images indicates material-related information being retained beyond range geometry alone.}
\label{fig:cluster_materials}
\end{figure*}

In \figref{fig:cluster_general_structure}, ceilings, floors, and walls occupy broad, internally consistent color regions whose transitions align with structural boundaries visible in the RGB reference and predicted range image.
Although the radar-predicted maps are smoother than the camera-derived maps, they preserve major distinctions among walls, doors, windows, and recessed passages that a coarse structural labeling would collapse.
Similar organization is visible at the object level: in \figref{fig:cluster_objects}, stairs, humans, and chairs appear as localized, coherent regions whose boundaries generally follow the predicted geometry.
This behavior appears across multiple examples of the same type of structure or object rather than in only one scene.
The examples in \figref{fig:cluster_materials} extend this observation to material-related regions: wood doors, a glass window, and a metal cabinet remain distinguishable from adjacent walls even when the surfaces are approximately coplanar.
However, these differences can arise from contextual associations learned from the camera-derived supervision.
The visualization alone does not separate these sources of information.

Taken together, these results provide a prompt-free semantic visualization of the learned feature organization, revealing how different scene regions separate or remain close in the embedding space without imposing predefined labels or clusters.
This complements conventional RF representations, which primarily emphasize reflection strength or geometry~\cite{yanik2020development,park2025diffusion_liliqiu,UCSD_3D_pc,dodds_mmnorm,panoradar,cartoradar,holoradar}, by exposing semantic transitions that are otherwise difficult to observe directly.
At the same time, differences between the radar-predicted and camera-derived projections indicate concepts for which the two sensing modalities organize the scene differently.
We examine these modality-dependent semantic differences more systematically in the following section (\secref{sec:capability}).

\subsection{Camera-to-radar concept transferability}
\label{sec:capability}
A shared 3D field can fuse repeated observations, but the meaning it retains still depends on the information predicted from radar.
To characterize the language evidence available to the field, we measure radar-camera agreement by comparing the text relevancy predicted from RF with that derived from the camera-based panoramic reference for the same prompts.
The camera-derived CLIP features serve as the supervisory reference rather than absolute semantic ground truth, so this analysis measures which camera-derived concept responses the distillation model reproduces after distillation.
Accordingly, it evaluates the features supplied to the field and does not measure a gain from scene-level fusion.

Our prompt bank contains 42 descriptive queries across five categories: objects or structures, colors, materials, human actions, and text-related concepts, as detailed in \secref{sec:query_analysis}.
The text-related prompts describe displayed content rather than the display itself, so recognizing the geometry of a sign does not count as recognizing its text.
To reduce confounding from object identity, we evaluate each non-geometric property within the entity that carries it, following \secref{sec:query_analysis}.
Specifically, actions are evaluated within person regions, colors and materials within one object class at a time, and text within display regions; objects and structures are evaluated over the full frame.
We report three complementary measures: pairwise discrimination for semantic preference, region AUROC for spatial localization, and instance-level correlation for variation across observations.
Chance performance is 0.50 for the first two measures, while the null reference for correlation is 0.

\begin{table*}[t]
\centering
\caption{\textbf{Per-building radar-camera concept agreement.} Each building is the held-out test split in the leave-one-building-out protocol.}
\label{tab:concept_capture_building}
\footnotesize
\begin{tabular*}{\textwidth}{@{\extracolsep{\fill}}lcccccccc@{}}
\toprule
Category & B1 & B2 & B3 & B4 & B5 & B6 & B7 & Mean \\
\midrule
\multicolumn{9}{l}{\emph{Pairwise discrimination} (chance 0.50)} \\
Objects / structures & 0.819 & 0.772 & 0.776 & 0.840 & 0.811 & 0.900 & 0.860 & 0.825 \\
Materials & 0.860 & 0.893 & 0.813 & 0.858 & 0.770 & 0.929 & 0.867 & 0.856 \\
Human actions & 0.741 & 0.741 & 0.708 & 0.777 & 0.733 & 0.787 & 0.799 & 0.755 \\
Colors & 0.636 & 0.609 & 0.563 & 0.651 & 0.643 & 0.790 & 0.793 & 0.669 \\
Text & 0.439 & 0.583 & 0.565 & 0.510 & 0.503 & 0.730 & 0.726 & 0.579 \\
\midrule
\multicolumn{9}{l}{\emph{Region AUROC} (chance 0.50)} \\
Objects / structures & 0.837 & 0.847 & 0.821 & 0.868 & 0.834 & 0.905 & 0.901 & 0.859 \\
Materials & 0.686 & 0.674 & 0.684 & 0.650 & 0.635 & 0.763 & 0.769 & 0.694 \\
Human actions & 0.683 & 0.642 & 0.702 & 0.674 & 0.648 & 0.803 & 0.778 & 0.704 \\
Colors & 0.552 & 0.561 & 0.553 & 0.573 & 0.550 & 0.682 & 0.670 & 0.592 \\
Text & 0.576 & 0.561 & 0.533 & 0.592 & 0.536 & 0.754 & 0.677 & 0.604 \\
\midrule
\multicolumn{9}{l}{\emph{Instance-level correlation} (null 0.00)} \\
Objects / structures & 0.847 & 0.891 & 0.823 & 0.926 & 0.851 & 0.886 & 0.970 & 0.885 \\
Materials & 0.507 & 0.382 & 0.542 & 0.449 & 0.279 & 0.807 & 0.827 & 0.542 \\
Human actions & 0.512 & 0.559 & 0.407 & 0.402 & 0.435 & 0.826 & 0.722 & 0.552 \\
Colors & 0.183 & 0.018 & 0.179 & 0.197 & 0.197 & 0.604 & 0.531 & 0.273 \\
Text & 0.067 & 0.180 & 0.153 & 0.266 & 0.175 & 0.684 & 0.418 & 0.277 \\
\bottomrule
\end{tabular*}
\end{table*}

\begin{figure*}[t]
\centering
\includegraphics[width=\textwidth]{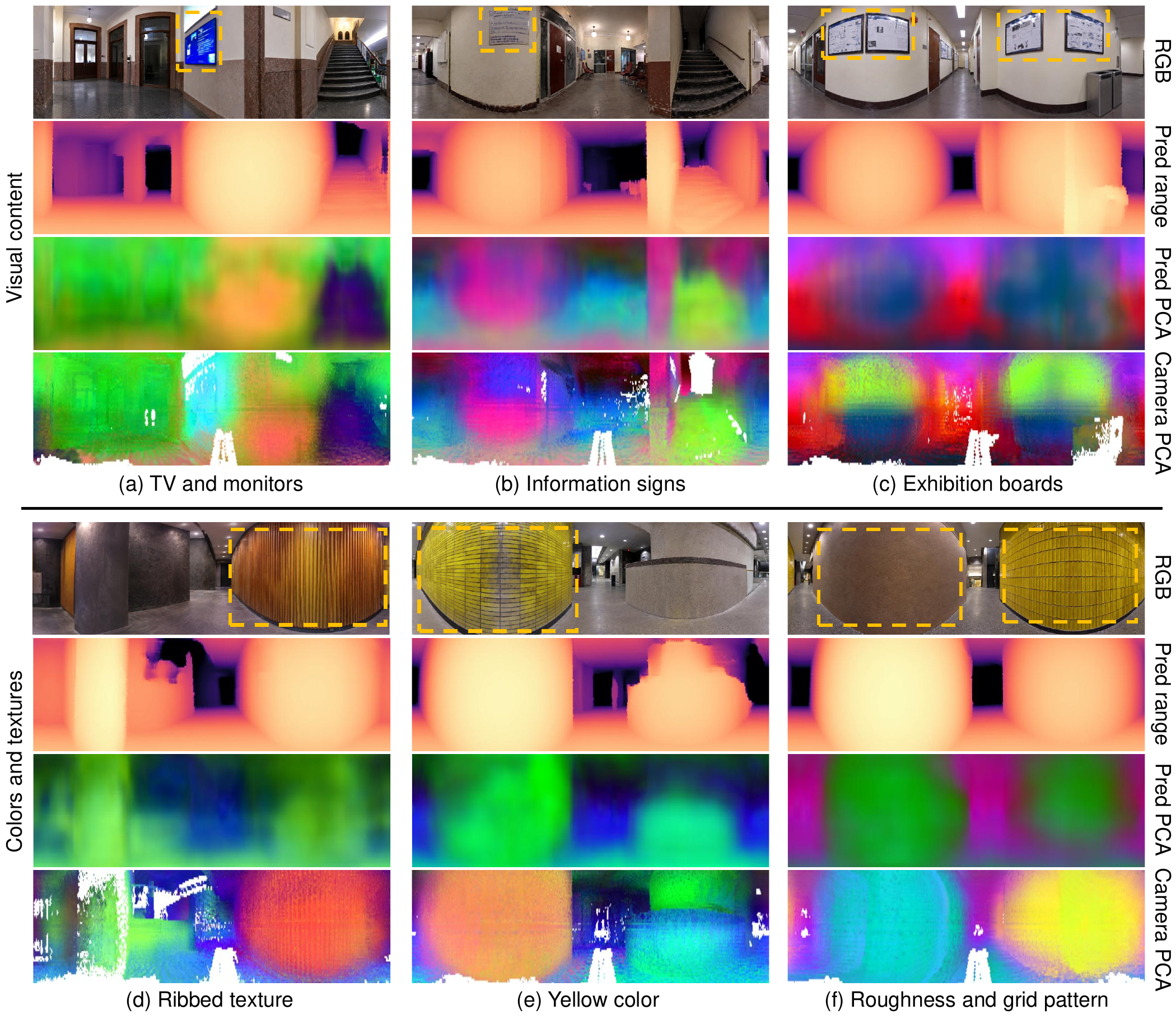}
\caption{\textbf{Radar-camera feature disagreement.}
The top row highlights concepts dominated by visual content: (a) televisions and monitors, (b) information signs, and (c) exhibition boards.
The bottom row highlights appearance-dependent concepts: (d) ribbed texture, (e) yellow color, and (f) roughness and grid patterns.
Dashed boxes mark the relevant regions in the RGB observations.
While the camera-derived features clearly distinguish these regions by displayed content, color, or fine surface appearance, the radar-predicted features often assign them similar embeddings and thus fail to distinguish them.}
\label{fig:radar_camera_disagree}
\end{figure*}

The category-wise results in \tabref{tab:concept_capture} show stronger radar-camera agreement for objects, structures, materials, and human configurations than for colors and printed text.
Objects and structures achieve pairwise discrimination, region AUROC, and instance-level correlation of 0.825, 0.859, and 0.885, respectively.
Materials and human actions also transfer substantially, whereas colors and printed text are much weaker, especially in instance-level correlation.
This category-level trend is broadly preserved across the seven held-out buildings in \tabref{tab:concept_capture_building}.
Because the categories use different prompts and evaluation regions, their absolute scores are not directly comparable as measures of intrinsic difficulty.
Nevertheless, the overall trend is consistent with mmWave sensitivity to geometry and material response~\cite{surfradar,shang2022liqray,xie2019tagtag}.
The human-action prompts primarily describe body poses visible within a radar panorama, so these results do not establish temporal action recognition.
In contrast, color and printed content are visual cues that camera-based methods can detect~\cite{scene_text_det}, but they have no general one-to-one relation to the measured RF response.
Their residual agreement may therefore come from correlations with object identity or surrounding context.

These possible contextual associations also suggest a closer look at material agreement.
Evaluating one object class at a time reduces instance-level correlation by 0.032 for materials and 0.041 for colors, confirming that object identity contributes to both.
However, the conditioned material correlation remains approximately twice that of color.
Together with the material queries in \figref{fig:single_frame_query} and the feature organization in \figref{fig:cluster_materials}, this supports material-sensitive semantics, but not direct estimation of physical material parameters.
The qualitative disagreements in \figref{fig:radar_camera_disagree} reinforce this boundary: camera-derived features distinguish displayed content, color, and fine surface appearance more clearly than radar-predicted features.
Overall, this analysis characterizes the concept evidence available to the occupancy-language field: strong agreement for scene structure, useful but context-dependent agreement for materials and human poses, and weaker agreement for visual appearance and text.

\subsection{Multiscale language-feature distillation}
\label{sec:clip_distillation}
The preceding results depend on the quality of the radar observations supplied to the scene field.
We therefore evaluate the distillation model separately, measuring its range accuracy and its ability to predict camera-derived language features in held-out buildings.
The model maps one panoramic RF volume to a dense range image and five spatially aligned CLIP feature maps, each with 768 channels.
The scales correspond to camera crop sizes of $[128,224,320,416,512]$ pixels and preserve different amounts of surrounding context.
We use a ViT-L-14-quickgelu OpenCLIP model~\cite{openclip} pretrained on DFN-2B~\cite{dfn2b} to construct the targets.
The architecture and geometry-aware supervision are described in \secref{sec:joint_frame}--\ref{sec:pan_gt}.

\begin{table*}[t]
\centering
\caption{\textbf{Per-building geometry and multiscale CLIP feature prediction in held-out environments.} Each building is evaluated as the held-out test split in the leave-one-building-out protocol.}
\label{tab:depth_clip}
\resizebox{\textwidth}{!}{%
\begin{tabular}{ccccccccc}
\toprule
 & \multicolumn{3}{c}{Range estimation} & \multicolumn{5}{c}{Radar-camera cosine similarity (\%)} \\
\cmidrule(lr){2-4}\cmidrule(lr){5-9}
Building & Median (cm) $\downarrow$ & Mean (cm) $\downarrow$ & PSNR $\uparrow$ & Scale 1 $\uparrow$ & Scale 2 $\uparrow$ & Scale 3 $\uparrow$ & Scale 4 $\uparrow$ & Scale 5 $\uparrow$ \\
\midrule
B1 & 5.49 & 17.92 & 26.16 & 92.13 & 92.62 & 92.56 & 91.86 & 91.12 \\
B2 & 5.00 & 19.77 & 26.36 & 90.76 & 91.56 & 91.71 & 91.07 & 90.19 \\
B3 & 5.50 & 24.04 & 23.16 & 88.59 & 89.20 & 88.06 & 88.44 & 87.62 \\
B4 & 4.03 & 15.16 & 28.93 & 93.42 & 93.67 & 93.57 & 92.93 & 91.94 \\
B5 & 4.49 & 14.64 & 29.04 & 91.06 & 91.39 & 91.14 & 90.15 & 89.03 \\
B6 & 3.65 & 11.89 & 28.51 & 94.28 & 94.33 & 94.25 & 93.82 & 93.13 \\
B7 & 3.08 & 11.25 & 29.52 & 94.83 & 94.70 & 94.36 & 93.76 & 93.96 \\
\bottomrule
\end{tabular}
}
\end{table*}

We evaluate the model along two axes: geometric accuracy and language-feature distillation quality.
For geometry, we report the median and mean absolute per-pixel range errors and the range-image PSNR.
For language-feature distillation, we compute cosine similarity between the radar-predicted feature and the camera-derived reference feature at every valid pixel and scale.
\tabref{tab:depth_clip} summarizes the leave-one-building-out results, where each building is used once as the held-out test environment.
Across all held-out buildings, the model achieves a mean range error of 16.38~cm, a mean PSNR of 27.38~dB, and a mean radar-camera cosine similarity of 91.92\%.
The population standard deviations across the seven building-level values are 4.20~cm, 2.12~dB, and 1.95\%, respectively.
The scale-wise values are similar across all five crop sizes.
Because an aggregate cosine score can remain high even when spatial localization is smooth, the spatial query maps in \secref{sec:scene_results} provide a complementary evaluation of local structure.

\begin{table*}[t]
\centering
\caption{\textbf{Model and decoder ablation for range and CLIP feature prediction.} We compare a transformer-based dense prediction model, a CNN with separate task decoders, and our CNN with a shared decoder. Metrics are averaged over all held-out buildings and, for cosine similarity, all five scales.}
\label{tab:model_decoder_ablation}
\resizebox{\columnwidth}{!}{%
\begin{tabular}{lccc}
\toprule
Model & Range error (cm) $\downarrow$ & Range image PSNR (dB) $\uparrow$ & Radar-camera cosine similarity (\%) $\uparrow$ \\
\midrule
DPT with shared decoder & 16.29 & 27.44 & 88.36  \\
UNet with separate decoder & 16.35 & 27.40 & 90.08 \\
UNet with shared decoder (ours) & 16.38 & 27.38 & 91.92 \\
\bottomrule
\end{tabular}
}
\end{table*}

To examine which architectural choices contribute to this feature agreement, we compare the distillation model with two architectural variants.
The first variant replaces the CNN backbone with the DPT transformer architecture for dense prediction~\cite{DPT}.
Our DPT variant uses 12 transformer layers with reassemble blocks placed after layers $[2,4,6,8,10,12]$, followed by a shared decoder and the same range and CLIP feature output heads as our distillation model.
The second variant keeps our CNN backbone but separates the range and CLIP branches at the bottleneck, using one decoder for range prediction and another for language-feature prediction.
\tabref{tab:model_decoder_ablation} summarizes the comparison.
The CNN and transformer models achieve similar range accuracy, but in our implementation the transformer gives weaker feature prediction and requires approximately twice the training time.
We hypothesize that dense RF language distillation benefits from the locality and multiscale feature hierarchy of the convolutional model, since semantic supervision is strongly tied to object boundaries, surfaces, and spatially localized scattering patterns.
A similar pattern appears in the separate-decoder CNN, which obtains comparable range accuracy but lower radar-camera cosine similarity.
This gap suggests that language-feature prediction benefits from sharing the decoder with range prediction, where both heads receive spatial features associated with geometric boundaries.
However, the similar range errors do not establish a geometric improvement from decoder sharing.

\subsection{Experimental setup}
\label{sec:setup}

\begin{figure*}[t]
\centering
\includegraphics[width=0.50\linewidth]{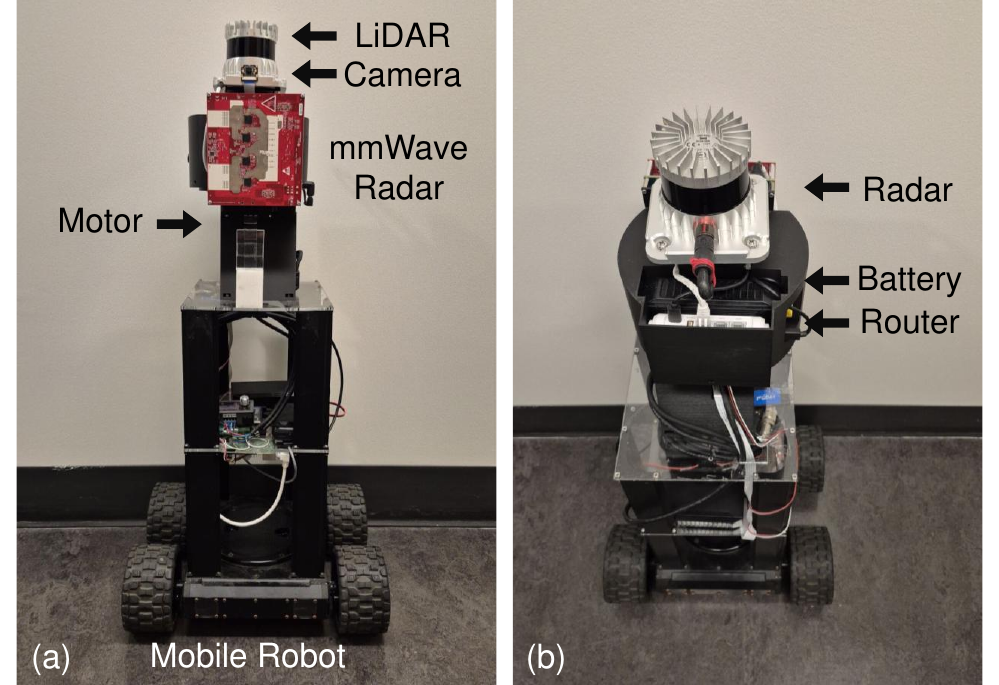}
\caption{\textbf{Hardware platform for data collection.}
The mobile robot carries a rotating mmWave MIMO radar, a 3D LiDAR, and a front-facing camera.
A stepper motor rotates the radar, while the battery and WiFi router rotate with the radar to prevent cable twisting and are placed behind it for weight balancing.
Radar commands are sent through the wireless router. (a) Front view. (b) Rear view.}
\label{fig:setup}
\end{figure*}

Our data collection platform follows the rotating-radar design of prior high-resolution RF imaging systems~\cite{panoradar,holoradar}.
The goal of this design is to convert a compact radar into a large synthetic aperture while keeping the sensor stack small enough for mobile indoor data collection.
As shown in \figref{fig:setup}, the platform uses a TI MMWCAS-RF-EVM cascaded mmWave radar and an MMWCAS-DSP-EVM data capture board to transmit and receive radio signals.
The radar consists of four cascaded chips with 9~TX and 16~RX channels enabled, forming a linear virtual array with 86 antenna elements after accounting for repeated elements.
A stepper motor rotates the radar at 2~Hz with a radius of 10.5~cm, producing a synthetic cylindrical array of size 86$\times$1024.
The radar transmits FMCW signals sweeping from 77 to 81~GHz, with a 4~GHz bandwidth and 256 ADC samples per chirp.
After beamforming, each panoramic RF observation is a tensor with 64 elevation bins, 512 azimuth bins, and 256 range bins.
To pair RF observations with geometry and language supervision, we mount an Ouster OS0-64 3D LiDAR and an IMX219 front-facing RGB camera on the same robot.
The LiDAR provides range supervision and geometric ground truth, while the camera provides images for generating multiscale language supervision.

\begin{figure*}[t]
\centering
\includegraphics[width=\linewidth]{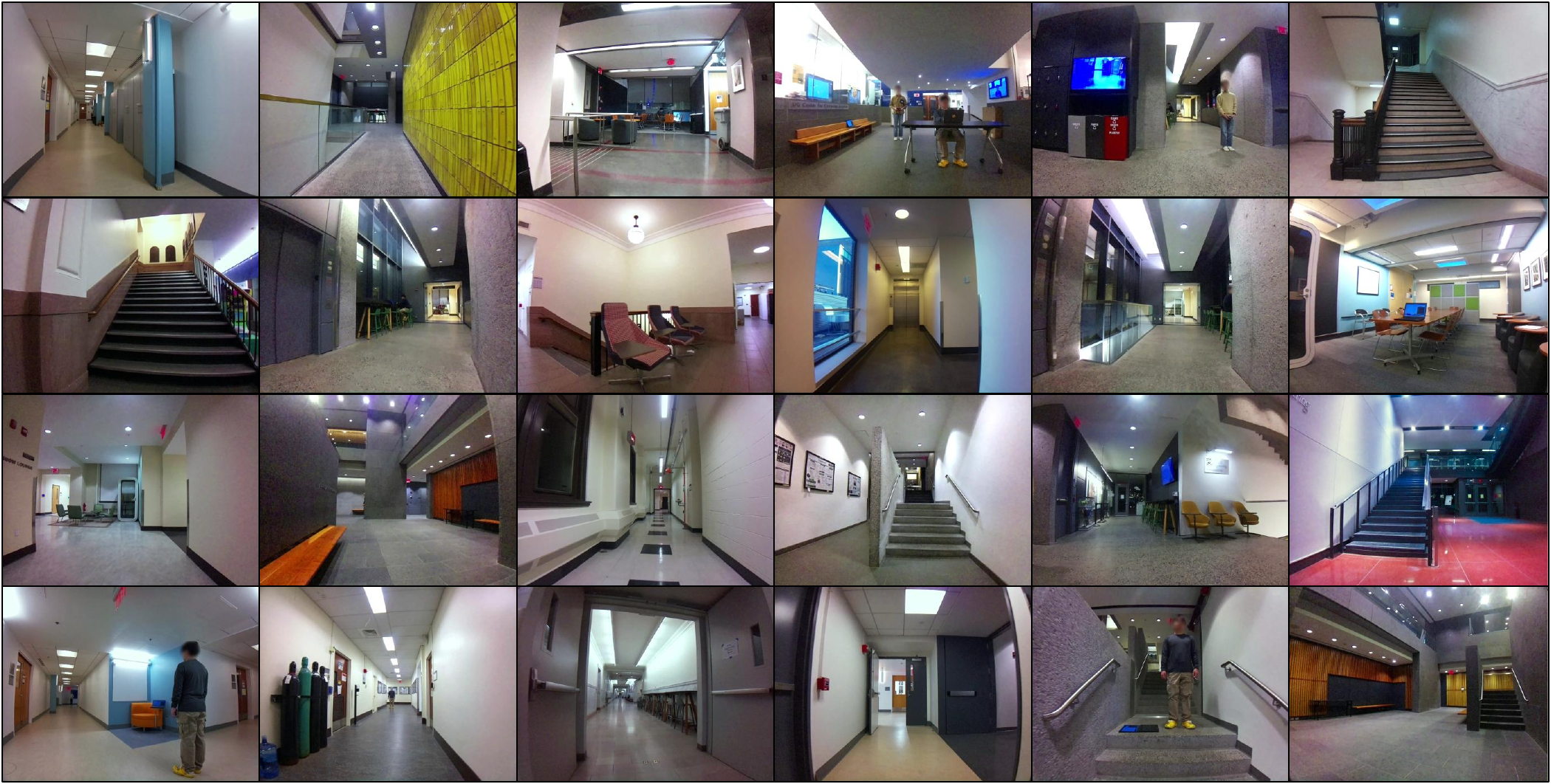}
\caption{\textbf{Example scenarios in our dataset.} The dataset spans common indoor environments, including hallways, lounges, offices, classrooms, and open floor areas. These examples are captured by the front-facing camera on the robot and illustrate the visual context used to generate language supervision.}
\label{fig:scene_example}
\end{figure*}

The resulting dataset contains 21 trajectories collected across seven buildings constructed between 1900 and 2010.
These buildings cover common indoor spaces such as offices, hallways, lounges, classrooms, and open floor areas, with representative examples shown in \figref{fig:scene_example}.
In total, the dataset includes approximately 20,000 synchronized radar-LiDAR-camera frames.
During data collection, the robot is manually controlled with a joystick and moves at a maximum speed of 0.6~m/s.
For each trajectory, we include at least one revisit to support loop closure and pose correction.
We also drive the robot through the same area in both directions so that the front-facing camera observes both sides of the environment over the course of a trajectory.

We evaluate generalization with a leave-one-building-out protocol.
\name{} contains two learned components with different training scopes.
For each split, we train the single-frame radar distillation model on trajectories from six buildings and apply it without adaptation to the held-out building.
We initialize its shared backbone and range head on the HoloRadar dataset~\cite{holoradar}, training for 90k iterations with AdamW, a momentum-based optimizer~\cite{momentum_freq}, a cosine learning-rate scheduler with base learning rate $10^{-4}$, and mixed precision.
We then fine-tune the full model on our collected radar-LiDAR-camera dataset for 60k iterations to refine range prediction and learn CLIP-aligned multiscale features.
This second stage uses the same optimizer and learning-rate schedule.
During training, we apply a random azimuth rotation to every sample and a random horizontal flip with probability 0.5.
Once the distillation model has been trained, we optimize the second component, the per-scene occupancy-language field, separately for each trajectory using the predicted ranges, predicted language features, and radar-estimated poses.
Our field uses these radar-derived inputs, where camera and LiDAR observations provide training supervision and evaluation references.
We use the same optimization settings for all evaluated trajectories.
All experiments are run on an AMD EPYC 7543 CPU and an NVIDIA L40S GPU.

\section{Discussion}
\label{sec:discussion}
\name{} makes a scene-level 3D occupancy-language field the interface between radar observations and spatial language queries.
Within this framework, the single-frame radar distillation model supplies surface and language predictions, while the field integrates them at shared 3D locations.
This distinction matters when a robot revisits a region or searches for a concept distributed across several views.
In either case, the language feature at a fixed scene location and context scale is independent of the observing viewpoint, and occupancy provides the geometry on which query responses are displayed.
The reconstructed scene can therefore be inspected with different prompts without refitting the field.

\begin{figure*}[t]
\centering
\includegraphics[width=\textwidth]{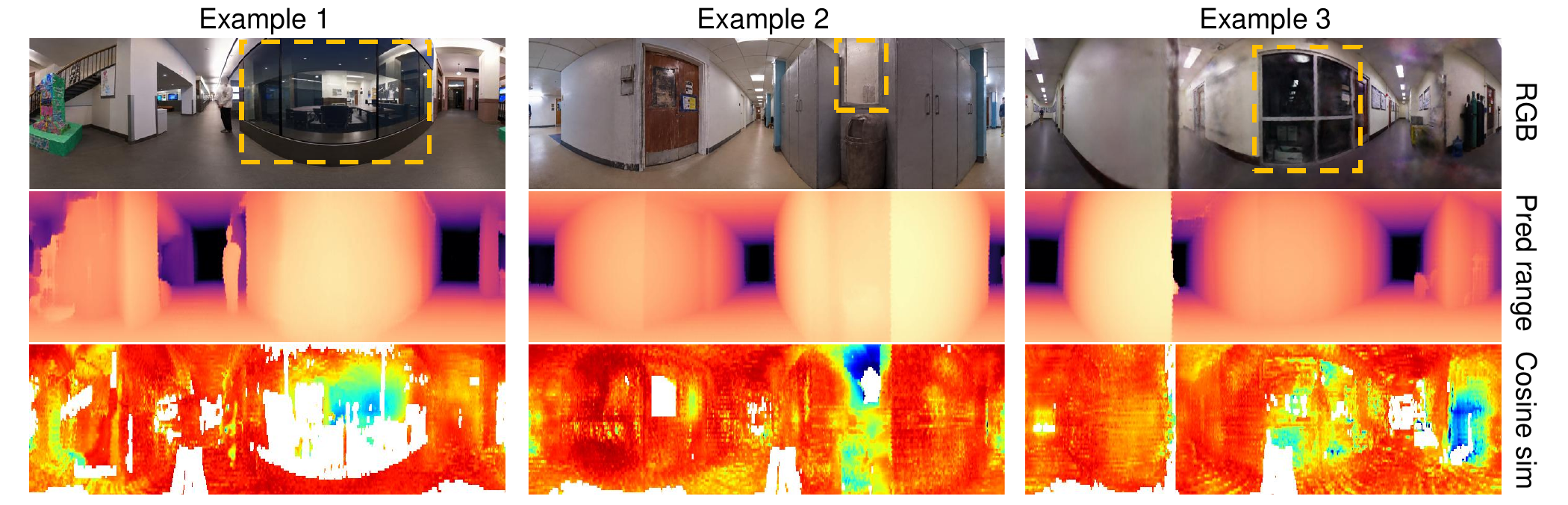}
\caption{\textbf{Radar-camera feature disagreement around glass.}
Each column presents one example, and the rows show the rendered panoramic RGB reference, the radar-predicted range image, and the pixelwise cosine similarity between radar-predicted and camera-derived CLIP features.
Dashed orange boxes indicate glass windows.
Warmer colors represent stronger feature agreement, and cooler colors represent weaker agreement.
Similarity decreases around the indicated glass regions, where the two sensors observe different scene content.}
\label{fig:glass_clip}
\end{figure*}

Interpreting the results also requires accounting for the physical differences between radar and cameras.
A low feature similarity may indicate a prediction error, but it can also arise because the sensors observe different aspects of the same region.
Glass provides a representative example.
In \figref{fig:glass_clip}, the camera records surface appearance, reflections, transmitted light, and objects visible beyond the pane, so its CLIP feature mixes several visual cues.
By contrast, many common glass surfaces behave as strong first reflectors for the radar configuration used here~\cite{glass_mmWave,panoradar}, and content visible behind them may not appear in the RF range image.
The resulting difference lowers radar-camera feature similarity around the windows.
Despite this lower agreement, the query in \figref{fig:single_frame_query} responds to the glass region, and the prompt-free visualization in \figref{fig:cluster_materials} separates it from the surrounding wall.
Radar features can therefore retain useful information about a surface without reproducing all of the visual content observed through it.

Beyond these modality differences, the results also clarify how to obtain useful observations for the field.
In particular, the distillation model benefits from coupling range and language-feature prediction.
Many feature transitions occur near object boundaries, planar surfaces, or changes in RF reflection behavior.
By preserving a common spatial feature hierarchy until the final task-specific heads, the shared decoder lets both outputs draw on these boundaries.
Consistent with this interpretation, separating the decoders changes mean range error by only 0.03~cm but reduces radar-camera cosine similarity from 91.92\% to 90.08\% in \tabref{tab:model_decoder_ablation}.
This supports decoder sharing for language-feature distillation, although it does not demonstrate a corresponding improvement in range accuracy.

Alongside the accuracy of individual observations, their spatial coverage matters for scene reconstruction.
Panoramic supervision determines the coverage of the language observations available along a trajectory.
During development, supervision restricted to the front-facing camera sector produced organized features in the repeatedly supervised region but poor features elsewhere.
Azimuth augmentation alone did not remove this behavior, as it changed sample coordinates without supplying the missing language targets.
Accumulation through robot motion addresses this limitation by providing valid targets across the panoramic grid, as detailed in \secref{sec:pan_gt}.

Several limitations remain in both the field and the distillation model.
First, the field describes a predominantly static scene.
Moving people and objects can create inconsistent surface and language observations unless dynamic regions are detected or modeled separately~\cite{dnerf}.
In our current system, people are treated as static scene elements, without persistent tracking or temporal action recognition.
Coupling the field with language-conditioned human motion models~\cite{mdm,moscale} is a possible route toward querying human movement over time.
Second, weak or unresolved reflections can cause compact objects to merge with neighboring surfaces, and shared 3D coordinates cannot recover a distinction absent from the observations.
Third, camera-derived features provide a supervisory reference rather than independent semantic ground truth.
Conditioning on object class reduces some confounds in the transferability analysis, but material, geometry, and context can remain correlated.
Further evaluation with human-annotated 3D query regions~\cite{Scanrefer}, controlled objects, and a geometry-only baseline would help distinguish these sources of agreement more directly.
Finally, dense 768-dimensional language observations impose substantial memory and storage costs.
A compact radar-specific feature representation could reduce these costs while retaining compatibility with text queries and occupancy-language field reconstruction.

\section{Methods}
\label{sec:method}
\begin{figure*}[t]
\centering
\includegraphics[width=0.62\textwidth]{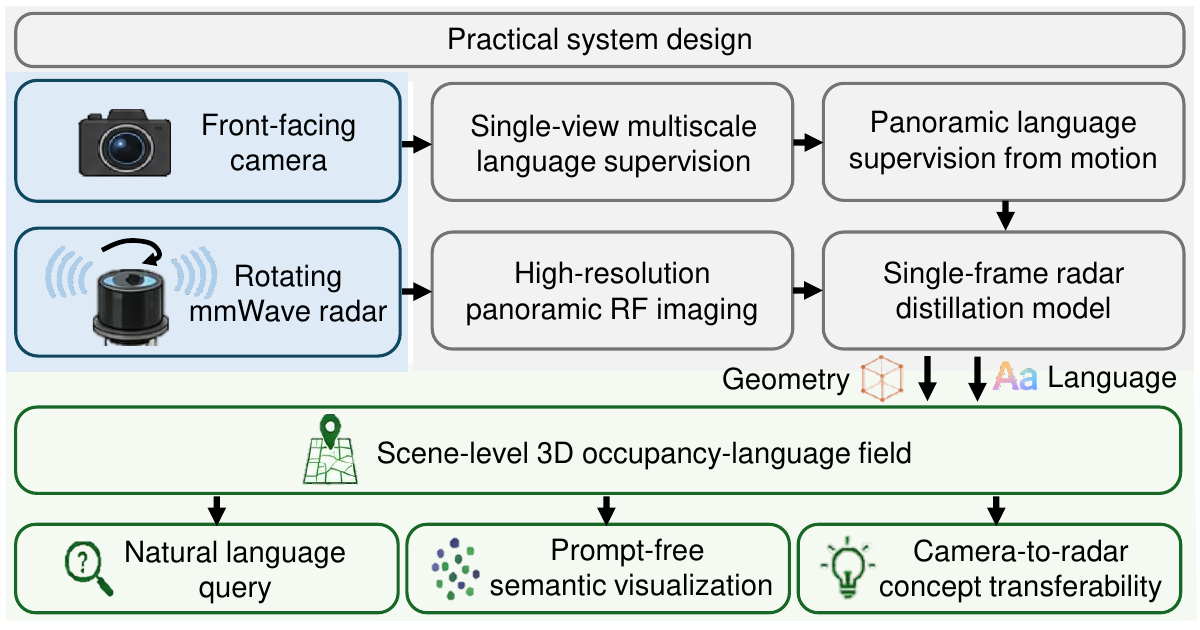}
\caption{\textbf{Overview of the \name{} framework.}
A rotating mmWave radar provides high-resolution panoramic RF measurements, while a front-facing camera supplies multiscale language supervision during data collection.
Building on these inputs, \name{} jointly learns geometry and language-aligned embedding from RF measurements and builds a scene-level 3D occupancy-language field, which supports several tasks.}
\label{fig:method_overview}
\end{figure*}

\name{} builds a scene-level 3D occupancy-language field from a sequence of radar-derived observations.
We first describe this field and how registered range and language predictions constrain its geometry and features.
We then describe the single-frame radar distillation model that supplies those predictions, together with its RF imaging input and camera-LiDAR supervision, as summarized in \figref{fig:method_overview}.
The final subsections specify the concept-transfer evaluation and deployment requirements.

\subsection{Scene-level 3D occupancy-language field}
\label{sec:lang_field}
The 3D occupancy-language field is the central component of \name{}.
Given panoramic radar observations $\{\mathbf{X}_k\}$ and their sensor poses $\{\mathbf{T}_k\}$, we reconstruct occupancy and language features in a common world coordinate system.
To provide the observations for this reconstruction, the distillation model supplies a range $\hat{D}_k(\mathbf{q})$ and normalized language features $\{\hat{\mathbf{E}}_{k,\ell}(\mathbf{q})\}_{\ell=1}^{L}$ for each elevation-azimuth pixel $\mathbf{q}$.
We describe this model in \secref{sec:joint_frame}.
Here, we focus on the field that integrates its outputs.
We define the occupancy-language field as
\begin{equation}
    \mathcal{F}(\mathbf{x},a)
    = \bigl(f_O(\mathbf{x}), f_L(\mathbf{x},a)\bigr),
    \qquad
    f_O(\mathbf{x})\in[0,1],
    \quad
    f_L(\mathbf{x},a)\in\mathbb{S}^{C-1},
    \label{eq:geometry_language_field}
\end{equation}
where $\mathbf{x}\in\mathbb{R}^3$ is a scene location, $a>0$ is a context-scale coordinate, and $\mathbb{S}^{C-1}$ denotes unit-norm language embeddings with $C=768$.
Occupancy describes the scene geometry, while the language field describes a location in relation to context at different scales.
Neither field takes the viewing direction as input.
Consequently, observations registered to the same location constrain a shared occupancy value, and observations at the same location and scale constrain a shared language embedding.

To implement these two components, we build on the occupancy mapping formulation of CartoRadar~\cite{cartoradar} and the scale-dependent language-field formulation of LERF~\cite{lerf}.
The resulting occupancy and language networks have separate parameters but share the scene coordinates and surface observations.
The occupancy network uses a multiresolution spatial grid and a small multilayer perceptron.
For language, we use a four-dimensional hash-grid encoding of normalized position and log scale, followed by a multilayer perceptron and $\ell_2$ normalization.
This parameterization makes the field continuous in position and scale while allowing geometry and language features to have different representational capacities.

\header{Occupancy from radar-predicted ranges.}
For pixel $\mathbf{q}$ in frame $k$, the pose defines a world-space sensor origin $\mathbf{o}_k$ and unit ray direction $\mathbf{d}_{k,\mathbf{q}}$.
The ray is $\mathbf{r}_{k,\mathbf{q}}(s)=\mathbf{o}_k+s\mathbf{d}_{k,\mathbf{q}}$, and the predicted range locates the surface at
\begin{equation}
    \hat{\mathbf{x}}_{k,\mathbf{q}}
    = \mathbf{o}_k+\hat{D}_k(\mathbf{q})\mathbf{d}_{k,\mathbf{q}}.
    \label{eq:field_surface_point}
\end{equation}
Using this surface estimate, we sample points along the ray within the reconstruction range and supervise occupancy using the predicted surface distance.
Samples in front of the surface receive free-space evidence, samples near the surface receive occupied-space evidence, and samples behind it receive little or no evidence from that observation.
To account for uncertainty in the surface location, we follow the probabilistic occupancy formulation of CartoRadar~\cite{cartoradar} and use a Laplace distribution centered on the predicted range.
If $F_D$ is its cumulative distribution function and $\epsilon_s$ is a surface tolerance, the free-space and surface weights at sample distance $s_m$ are proportional to $1-F_D(s_m+\epsilon_s)$ and $F_D(s_m+\epsilon_s)-F_D(s_m-\epsilon_s)$, respectively.
The resulting objective is a weighted binary cross-entropy,
\begin{equation}
    \mathcal{L}_{\mathrm{occ}}
    = \frac{1}{2M}\sum_{m=1}^{M}
    \left[
        w_m^{\mathrm{free}}\,\mathrm{BCE}(O_m,0)
        + w_m^{\mathrm{surf}}\,\mathrm{BCE}(O_m,1)
    \right],
    \quad
    O_m=f_O(\mathbf{r}(s_m)),
    \label{eq:field_occupancy_loss}
\end{equation}
where $M$ is the number of sampled points and the surface weight compensates for the much smaller spatial support of occupied observations.
This supervision fits occupancy directly at sampled points without rendering a depth image during field training.
Repeated rays from different poses constrain the same occupancy field and thereby fuse their geometric observations.

\header{Language features anchored to the surfaces.}
To attach language features to this geometry, we train the language field at the surface points in \eqref{eq:field_surface_point}, using language features predicted from radar.
To provide continuous supervision between the $L=5$ output scales, we select adjacent scales $\ell$ and $\ell+1$ and an interpolation weight $u\in[0,1]$.
The target embedding is
\begin{equation}
    \bar{\mathbf{e}}_{k,\mathbf{q},u}
    = \operatorname{norm}\!\left[
        (1-u)\hat{\mathbf{E}}_{k,\ell}(\mathbf{q})
        +u\hat{\mathbf{E}}_{k,\ell+1}(\mathbf{q})
    \right],
    \label{eq:field_scale_interpolation}
\end{equation}
where $\operatorname{norm}$ denotes $\ell_2$ normalization.
We interpolate the associated crop-scale parameters in the same manner, $p_u=(1-u)p_\ell+up_{\ell+1}$, and use the range-dependent scale coordinate $a=p_u\hat{D}_k(\mathbf{q})/f_s$.
Here $f_s$ is the configured conversion factor for the scene field.
The crop-support calibration used to construct the camera supervision is described in \secref{sec:point_lang_gt}.
Conditioning on this scale coordinate prevents features with different amounts of context from being forced into a single vector at each location.
For valid observations $\mathcal{V}$, the language objective is
\begin{equation}
    \mathcal{L}_{\mathrm{field\text{-}lang}}
    = \frac{1}{|\mathcal{V}|}
      \sum_{(k,\mathbf{q},u)\in\mathcal{V}}
      \sum_{c=1}^{C}
      h_\delta\!\left(
          [f_L(\hat{\mathbf{x}}_{k,\mathbf{q}},a)]_c
          -[\bar{\mathbf{e}}_{k,\mathbf{q},u}]_c
      \right),
    \label{eq:field_language_loss}
\end{equation}
where $h_\delta$ is the Huber loss and $\delta=1.25$.
Combining the geometry and language terms gives the field objective $\mathcal{L}_{\mathrm{scene}}=\mathcal{L}_{\mathrm{occ}}+\lambda_s\mathcal{L}_{\mathrm{field\text{-}lang}}$, with $\lambda_s=0.1$.
The geometry and language terms can use separate observation batches while optimizing their respective networks in the same scene frame.
Unlike LERF's language supervision through volume rendering, this formulation supervises the language field directly at estimated surface points.
Camera-derived embeddings are used to train the distillation model, but radar-predicted embeddings supply the language targets for deployment-time field construction.

\header{Querying and viewing the reconstructed scene.}
After fitting the field, we extract surface points from occupancy and evaluate language features at those locations.
A normalized CLIP text embedding $\mathbf{y}_t$ defines the pointwise query score $s_t(\mathbf{x},a)=f_L(\mathbf{x},a)^{\mathsf T}\mathbf{y}_t$.
With this score, the field supports querying at a selected scale or searching over a set of scales to obtain $s_t^\star(\mathbf{x})=\max_{a\in\mathcal{A}}s_t(\mathbf{x},a)$.
To visualize the response, we color surface points by their query scores.
The resulting maps retain the spatial extent and relative locations of retrieved regions within the reconstructed scene.

In addition to querying surface points directly, we can inspect the field by rendering depth and language features along rays from a chosen viewpoint.
The rendered depth is the distance of the sample with the largest termination weight.
The same geometry thus determines the visible surface and the language features accumulated along a ray.
Note that rendering is a way to inspect the fitted field but not the language-training objective in \eqref{eq:field_language_loss}.
For prompt-free visualization, language features at reconstructed surfaces can instead be projected to three PCA components and mapped to RGB.
Keeping this projection fixed across the scene gives each location a consistent displayed color across viewpoints.

\subsection{Single-frame radar distillation model}
\label{sec:joint_frame}
To supply the field with observations at each sensor pose, we train a single-frame radar distillation model on paired geometric and language targets.
The model learns ranges from LiDAR supervision and distills language features from a pretrained camera encoder.
Given a panoramic RF tensor $\mathbf{X}$, the model $f_{\boldsymbol{\eta}}$ jointly predicts a dense range image $\hat{\mathbf{D}}$ and $L=5$ multiscale language-feature maps $\{\hat{\mathbf{Z}}_{\ell}\}_{\ell=1}^{L}$.
We use a convolutional UNet-style architecture with a shared encoder and decoder.
The range and language tasks share the complete encoder-decoder backbone and separate only at the output heads, with one scalar range head and one $C$-dimensional language head for each scale, where $C=768$.
This design gives both outputs a common hierarchy of spatial features before their final prediction heads.

Before the encoder, a 7$\times$7 convolutional stem reduces the input range channels from 256 to 64.
This stem captures spatial energy patterns across elevation and azimuth while compressing the 256-channel range dimension of the RF tensor.
Within this backbone, both the encoder and decoder contain seven layers linked by skip connections.
Each encoder layer has four residual blocks, and each decoder layer has two residual blocks.
Feature maps are downsampled by a factor of two at the end of each encoder layer and upsampled correspondingly in the decoder.
To preserve elevation resolution (initially $N_\phi=64$), we stop downsampling along the elevation dimension after the fourth layer.
After the shared backbone extracts RF features from $\mathbf{X}$, task-specific heads produce the final outputs.
The range head contains two additional residual blocks, and each scale has a dedicated head with five additional residual blocks.

Because CLIP represents semantic similarity through embedding direction, we $\ell_2$-normalize each predicted language vector $\hat{\mathbf{Z}}_{\ell}(\mathbf{q})$ to obtain $\hat{\mathbf{E}}_{\ell}(\mathbf{q})$, where $\mathbf{q}$ denotes an elevation-azimuth pixel.
The camera-derived target $\mathbf{E}_{\ell}(\mathbf{q})$ is normalized in the same manner during panoramic supervision construction.

For geometric supervision, we use the LiDAR range image $\mathbf{D}$.
Let $\Omega_r$ denote the set of pixels with valid geometry targets.
The range loss $\mathcal{L}_{\mathrm{range}}$ is the mean Smooth L1 error between $\hat{\mathbf{D}}$ and $\mathbf{D}$ over $\Omega_r$.
The Smooth L1 loss provides robustness to isolated LiDAR noise while retaining sensitivity to small range errors.
We additionally compute an LPIPS loss $\mathcal{L}_{\mathrm{perc}}$ between the predicted and target range images after applying the same normalization and validity mask.
This term complements the pixel-wise range loss by encouraging coherent surfaces and sharper structural boundaries.
Alongside these geometric losses, the language objective maximizes cosine similarity between the predicted and camera-derived features at every valid scale and pixel.
Let $\Omega_{e,\ell}$ denote the validity set produced by the panoramic supervision procedure for scale $\ell$.
The multiscale embedding loss is
\begin{equation}
    \mathcal{L}_{\mathrm{embed}}
    =
    \frac{
        1
    }{
        \displaystyle
        \sum_{\ell=1}^{L}
        |\Omega_{e,\ell}|
    }
    \sum_{\ell=1}^{L}
    \sum_{\mathbf{q}\in\Omega_{e,\ell}}
    \left[
        1
        -
        \hat{\mathbf{E}}_{\ell}(\mathbf{q})^{\mathsf T}
        \mathbf{E}_{\ell}(\mathbf{q})
    \right].
    \label{eq:frame_embedding_loss}
\end{equation}
The scale-specific masks allow supervision density to vary across scales without treating missing language targets as negative examples.
Combining these losses gives the complete single-frame objective
\begin{equation}
    \mathcal{L}_{\mathrm{frame}}
    =
    \lambda_r\mathcal{L}_{\mathrm{range}}
    +
    \lambda_p\mathcal{L}_{\mathrm{perc}}
    +
    \lambda_e\mathcal{L}_{\mathrm{embed}},
    \label{eq:frame_total_loss}
\end{equation}
where we use
$\lambda_r=1.0$, $\lambda_p=0.1$, and $\lambda_e=0.5$.

The geometry and language validity masks are handled independently.
In particular, regions where LiDAR does not provide a reliable first-surface range, including problematic measurements around transparent surfaces, are excluded from the geometry losses.
At the same time, a language target is retained whenever it remains valid under the panoramic language-supervision procedure.
The remaining optimization protocol, model initialization, and data augmentation are described in \secref{sec:setup}.

For evaluation, the absolute range error is computed over the valid LiDAR pixels and summarized by its median and mean.
We compute range-image PSNR from the masked mean squared error using $D_{\max}=9.6$~m as the maximum signal value.
The reported radar-camera feature agreement is the mean cosine similarity between normalized predicted and reference features over valid pixels.

\subsection{High-resolution panoramic RF imaging}
\label{sec:rf_imaging}
Our learning pipeline operates on a dense panoramic RF volume rather than the sparse point clouds conventionally produced by a single-chip radar.
To obtain this volume, we adopt a motion-compensated cylindrical synthetic-aperture imaging procedure based on PanoRadar~\cite{panoradar}.
As the radar rotates, each physical antenna samples the environment from a sequence of positions around the rotation axis, forming a dense virtual cylindrical array.
This enlarged aperture substantially improves azimuth resolution while preserving the full $360^\circ$ field of view required for panoramic perception.

Let $t$ denote slow time during one radar rotation, and let $m$ index the virtual antenna elements along the vertical array.
The antenna position is $\mathbf{p}_{m}(t)=\mathbf{c}(t)+[\rho\cos(\omega t),\rho\sin(\omega t),z_m]^{\mathsf T}$, where $\rho$ and $\omega$ are the rotation radius and angular velocity, $z_m$ is the vertical antenna offset, and $\mathbf{c}(t)$ is the translational motion of the platform.
For a queried azimuth $\theta$ and elevation $\phi$, the unit direction is $\mathbf{u}(\theta,\phi)=[\cos\phi\cos\theta,\cos\phi\sin\theta,\sin\phi]^{\mathsf T}$.
The measurements are coherently combined by compensating for the round-trip propagation-phase difference associated with each virtual antenna position:
\begin{equation}
    y(\theta,\phi,\tau)
    =
    \sum_{m}
    \sum_{t\in\mathcal{W}(\theta)}
    w_{m,t}\,
    s_m(t,\tau)
    \exp\left(
        j\frac{4\pi}{\lambda}
        \mathbf{p}_{m}(t)^{\mathsf T}
        \mathbf{u}(\theta,\phi)
    \right),
    \label{eq:cylindrical_beamforming}
\end{equation}
where $s_m(t,\tau)$ is the complex FMCW intermediate-frequency signal at fast-time index $\tau$, $\lambda$ is the wavelength, and $w_{m,t}$ is the beamforming weight.
Because the physical antennas have a directional radiation pattern, $\mathcal{W}(\theta)$ includes only measurements within a window centered around the queried azimuth.
A range Fourier transform is subsequently applied along $\tau$, and its magnitude forms the beamformed reflection volume $X(\theta,\phi,r)$.

For this coherent combination to succeed, the antenna trajectory must be known accurately.
Ignoring platform motion causes phase inconsistencies among measurements collected at different rotation angles, which spatially shifts and blurs the combined image.
To compensate for these effects, we estimate the planar platform motion directly from the radar measurements rather than relying on an external odometry source.
After compensating for the known motor-driven rotation, we apply a sliding slow-time Fourier transform to isolate the angular and Doppler responses of static environmental reflectors.
For a reflector observed at azimuth $\theta_i$, its measured radial velocity follows $v_{r,i}=v\cos(\theta_i-\psi)$, where $v$ and $\psi$ denote the platform speed and heading direction over one rotation.
We estimate these motion parameters by collecting observations from multiple reflectors at different ranges and azimuths and robustly fitting the sinusoid using RANSAC.
The resulting motion estimate determines $\mathbf{c}(t)$, allowing the translational displacement to be incorporated directly into the phase compensation in \eqref{eq:cylindrical_beamforming}.
We implement the imaging operation as consecutive 1D azimuth and elevation beamforming steps, separating the corresponding steering terms to reduce computation.

Evaluating the beamforming and range transform over a panoramic grid produces a dense tensor $\mathbf{X} \in \mathbb{R}^{N_{\phi}\times N_{\theta}\times N_r}$, whose axes correspond to elevation, azimuth, and range.
This dense RF tensor preserves weak reflections and spatial energy spreading that would otherwise be discarded by conventional point-cloud extraction.
In the distillation model, we treat the range dimension as the input channels of a panoramic 2D convolutional network.
This design processes the panoramic RF observations efficiently before their predicted surfaces are registered in 3D.

\subsection{Single-view multiscale language supervision}
\label{sec:point_lang_gt}
To pair the RF volumes with language targets, we derive supervision from the synchronized front-facing RGB camera using a pretrained CLIP image encoder~\cite{openclip}.
Because CLIP produces an embedding for an image region rather than an individual pixel, we construct an overlapping multiscale crop pyramid following the general strategy of LERF~\cite{lerf}.
For each RGB frame, we extract square crops at five pixel scales, $p_\ell\in\{128,224,320,416,512\}$, using a stride of $p_\ell/2$.
The crops are encoded independently and $\ell_2$-normalized to form five spatial grids of CLIP features.
Smaller crops preserve localized objects and structures, whereas larger crops encode broader structural and contextual semantics.

Associating these image-region features with radar-observed surfaces also requires depth, which the front-facing RGB camera does not directly measure.
We therefore use the synchronized 3D LiDAR to associate the image embeddings with physical scene points.
For a LiDAR point $\mathbf{p}_i$, we transform it into the camera coordinate system as $\mathbf{x}^{c}_i=\mathbf{T}_{c\leftarrow l}\mathbf{p}_i$ and project it to image location $\mathbf{u}_i=\pi(\mathbf{x}^{c}_i)$.
Here, $\mathbf{T}_{c\leftarrow l}$ denotes the calibrated LiDAR-to-camera transformation and $\pi(\cdot)$ is the camera projection function.
We retain points with positive camera depth whose projections fall within the valid image region.
The camera and LiDAR are mounted with a small spatial baseline, which reduces parallax and visibility discrepancies between their observations, although minor mismatches can remain near occlusion boundaries.
Once these correspondences are established, we bilinearly interpolate the embeddings of the four neighboring crop centers for each valid point and crop scale, then normalize them.
This yields five camera-derived language targets for each camera-visible LiDAR point, rather than a dense pixel-wise language map over the panorama.

Because a fixed image crop covers a larger physical region at greater depth, we associate each feature $\mathbf{e}_{i,\ell}$ with an approximate physical scale
\begin{equation}
    a_{i,\ell}
    =
    \frac{p_\ell z_i}{f_{\mathrm{eff}}},
    \qquad
    f_{\mathrm{eff}}=\sqrt{f_xf_y},
\end{equation}
where $z_i$ is the point depth in the camera coordinate system and $f_x$ and $f_y$ are the camera focal lengths.
Each visible LiDAR point is consequently stored with its 3D position and the five embedding-scale pairs
$\{(\mathbf{e}_{i,\ell},a_{i,\ell})\}_{\ell=1}^{5}$.
These sparse, point-aligned language features are subsequently accumulated and projected to construct panoramic supervision for the RF measurements.

\subsection{Panoramic language supervision from robot motion}
\label{sec:pan_gt}
The front-facing camera provides language supervision over only a limited azimuthal region, whereas the RF input spans the full $360^\circ$ panorama.
As a result, supervision restricted to the camera field of view introduces a directional bias: the model receives no language targets for other parts of the same RF panorama.
Although stitching camera frames into RGB panoramas could extend this coverage, it is unreliable under robot vibration, turning motion, parallax, and accumulated pose error.
We instead exploit robot motion and LiDAR geometry to accumulate the sparse front-view language features in a global map and reproject them into each panoramic LiDAR frame.

To construct this map, we first estimate the LiDAR poses using a SLAM pipeline consisting of LiDAR odometry and loop-closure correction, following the strategy used in CartoRadar~\cite{cartoradar}.
For each frame, the camera-visible LiDAR points and their multiscale language features from \secref{sec:point_lang_gt} are transformed into the world coordinate system using the estimated pose.
Accumulating these observations over a trajectory produces a point-reference language map $\mathcal{M}$.
Each map entry contains a world-space LiDAR point $\mathbf{x}_n$ and its normalized CLIP feature $\mathbf{e}_{n,\ell}$ with physical support $a_{n,\ell}$ at every scale $\ell$.
Repeated observations of the same scene region are retained as separate references and are combined when the map is queried.

To generate panoramic supervision for frame $k$, we convert its panoramic LiDAR range image into points and transform every valid point $\mathbf{p}_{k,j}$ into the world frame as $\mathbf{q}_{k,j}=\mathbf{T}_{k}\mathbf{p}_{k,j}$, where $\mathbf{T}_{k}$ is the estimated LiDAR pose.
For each query point and scale $\ell$, we compute the desired physical support as $\hat{a}_{k,j,\ell}=p_\ell r_{k,j}/f_{\mathrm{eff}}$, using the query range $r_{k,j}$.
However, each map point stores language-aligned features at several physical scales determined by the depth at which it was originally observed.
The desired support $\hat{a}_{k,j,\ell}$ may therefore fall between two stored scales.
To match this support at a map point $\mathbf{x}_n$, we linearly interpolate the embeddings at the two neighboring physical scales to obtain
$\mathbf{e}_{n}(\hat{a}_{k,j,\ell})$, followed by $\ell_2$ normalization.
A map point is considered valid for the query only when the desired support lies within its stored scale range, up to a small tolerance.
We then retrieve the $K$ nearest map points to $\mathbf{q}_{k,j}$ within a maximum spatial radius $r_{\max}$.
Their scale-matched embeddings are aggregated using inverse-square spatial weights:
\begin{equation}
    \tilde{\mathbf{e}}_{k,j,\ell}
    =
    \frac{
        \displaystyle
        \sum_{n\in\mathcal{N}(\mathbf{q}_{k,j})}
        w_{kjn}\,
        \mathbf{e}_{n}(\hat{a}_{k,j,\ell})
    }{
        \displaystyle
        \sum_{n\in\mathcal{N}(\mathbf{q}_{k,j})}
        w_{kjn}
    },
    \qquad
    w_{kjn}
    =
    \frac{1}{
        \max\!\left(
            \lVert\mathbf{q}_{k,j}-\mathbf{x}_n\rVert_2,
            \epsilon
        \right)^2
    },
    \label{eq:panoramic_clip_aggregation}
\end{equation}
where $\mathcal{N}(\mathbf{q}_{k,j})$ contains only spatially valid neighbors with compatible scale embeddings and $\epsilon>0$ is a numerical stability constant.
We finally $\ell_2$-normalize the aggregated feature to obtain the panoramic language target $\mathbf{e}^{\mathrm{pano}}_{k,j,\ell}$.

A language target is marked valid when the query pixel has a corresponding panoramic LiDAR point and at least one nearby map point provides a compatible scale embedding.
Repeating this procedure over all LiDAR pixels and scales produces multiscale panoramic language targets and their corresponding validity masks.
This transfers language information collected by a single frontal camera to the full RF field of view while maintaining alignment with the panoramic LiDAR geometry.

\subsection{Concept transferability}
\label{sec:query_analysis}
We evaluate the language observations supplied to the field by comparing the single-frame radar predictions with camera-derived reference features.
This camera-to-radar agreement characterizes the concept evidence available before scene fusion.
Specifically, we apply the same prompt bank to radar-predicted and camera-derived features at the same pixels and quantify how closely their responses agree.
Let $s\in\{\mathrm{rad},\mathrm{cam}\}$ denote the feature source, $\mathbf{F}_s(\mathbf{q})$ the normalized feature at pixel $\mathbf{q}$ and a given scale, and $\mathbf{y}_t$ the normalized CLIP text embedding of prompt $t$.
However, raw cosine similarities occupy a narrow, prompt-dependent range and are therefore difficult to compare directly.
Following LERF~\cite{lerf}, we place them on a common relevancy scale by comparing each query with the fixed set of canonical negatives $\mathcal{C}=\{\texttt{object},\texttt{things},\texttt{stuff},\texttt{texture}\}$:
\begin{equation}
    r_s(\mathbf{q},t)
    =
    \left[
        1
        +
        \exp\left(
            -\beta
            \left[
                \mathbf{F}_s(\mathbf{q})^{\mathsf T}\mathbf{y}_t
                -
                \max_{c\in\mathcal{C}}
                \mathbf{F}_s(\mathbf{q})^{\mathsf T}\mathbf{y}_c
            \right]
        \right)
    \right]^{-1},
    \label{eq:lerf_relevancy}
\end{equation}
where $\mathbf{y}_c$ is the normalized text embedding of negative prompt $c$ and $\beta=10$ controls the sharpness of the mapping.
A value above $0.5$ means that prompt $t$ is more similar to the feature than the most similar canonical negative.
Setting $s=\mathrm{rad}$ or $s=\mathrm{cam}$ gives the radar and camera relevancies, respectively.
Because camera-derived features exist only at LiDAR-supervised pixels, these pixels define the valid set for every comparison.
The 42 descriptive queries spanning the five categories are listed in \tabref{tab:prompt_bank}.

\begin{table*}[t]
\centering
\caption{\textbf{Prompt bank for the camera-to-radar concept transferability analysis.} The bank contains 42 queries across the five semantic categories used in the evaluation.}
\label{tab:prompt_bank}
\resizebox{\textwidth}{!}{%
\begin{tabular}{p{1.5in} p{5in}}
\toprule
Category & Prompts \\
\midrule
Objects / structures (16) & a flat interior wall; the floor of the room; the ceiling overhead; a door in the hallway; a trash bin on the floor; a person standing in the room; a glass window or panel; a table; a chair; a staircase with steps; a vertical pillar or column; a handrail or railing; a television screen on the wall; a window on the wall; a carpeted floor; an elevator door \\
\addlinespace
Colors (9) & a \{red, blue, green, yellow, orange, black, white, gray, pink\} colored object \\
\addlinespace
Materials (6) & a wooden surface; a metal surface; a marble surface; an exposed aggregate concrete surface; a plastic surface; a ceramic surface \\
\addlinespace
Human actions (6) & a person standing upright; a person sitting down; a person lying on the floor; a person raising their arms; a person bending down; a person crouching \\
\addlinespace
Text (5) & a digital display showing numbers; printed letters on a sign; the number 7; the letter A; a line of printed words \\
\bottomrule
\end{tabular}
}
\end{table*}

Most semantic properties co-occur with particular objects, so we evaluate each non-geometric property within the entity that carries it.
For this conditioning, the camera-derived object prompts assign each valid pixel to its most relevant object class, and we retain assignments with relevancy above $0.5$.
Colors and materials are evaluated separately within trash-bin, table, chair, and display regions.
Similarly, text is evaluated within displays and human actions within person regions.
Objects and structures, in contrast, are evaluated over the full frame because their geometry is part of the concept.
Each frame, feature scale, and conditioning class defines one entity unit, and units with fewer than 20 valid pixels are discarded.
For each prompt, the camera-derived concept region is the top 5\% of camera-relevancy values within the unit.

We compare radar and camera responses using three complementary measures.
First, pairwise discrimination considers every prompt pair within a category and the pixels that belong to exactly one of their camera-derived concept regions.
It measures how often radar prefers the same prompt as the camera, pooling pixels across entity units before averaging equally over prompt pairs.
Its chance level is $0.5$.
Second, region AUROC treats the camera-derived concept region as positive and the remaining pixels in the same unit as negative.
It measures whether radar assigns higher relevancy to the concept region and has a chance level of $0.5$.
Scores are computed within each unit and then averaged across units and prompts.
Finally, the measure termed instance-level correlation first averages radar and camera relevancy within each entity unit, then computes their Pearson correlation across observations assigned to the same object class.
An entity unit is a frame--scale--class region; it does not require tracking a physical object instance across frames.
For this measure, a class must contain at least 10 observations, and correlations are averaged over prompts and eligible classes.
Its null reference is $0$.
All three measures include the five CLIP scales and are summarized by category and held-out building.

\subsection{Practical system design}
\label{sec:system_design}
\name{} separates training the single-frame radar distillation model from reconstructing a new scene.
The camera and LiDAR provide supervision for the first stage: camera features supply language targets, and LiDAR geometry supports range targets, cross-modal alignment, and the reference trajectory used to accumulate panoramic supervision.
Once trained, the distillation model predicts both ranges and language features from RF measurements.
It is reused across environments, while a separate occupancy-language field is fitted to the observations of each new scene.

Scene reconstruction also requires knowing sensor poses for radar predictions.
For the reported scene-level results, these poses are estimated from radar.
Radar odometry provides an initial trajectory, and geometric registration and loop-closure correction can reduce accumulated drift following the mapping strategy of CartoRadar~\cite{cartoradar}.
The occupancy implementation also supports pose refinement through its geometric reconstruction loss.
The language loss, by contrast, directly supervises surface features and does not itself optimize the sensor poses.
Nevertheless, both fields depend on the correspondence between observations and scene locations, so trajectory quality affects the consistency of the resulting geometry and language features.

To keep these training and reconstruction stages practical, the pipeline reuses learned models and extracted features to limit storage and computation.
During training-data construction, the panoramic reference map stores links to camera features and their spatial and scale information rather than duplicating every high-dimensional feature at every accumulated point.
During scene reconstruction, occupancy samples and language surface observations can be drawn separately, allowing each language batch to reuse features from one cached radar frame.
Finally, direct point supervision avoids volume rendering in the field-training losses, while occupancy-based rendering remains available for inspecting the fitted scene.

\section*{Declarations}

\header{Data availability} \newline
The data supporting the findings of this study are available from the corresponding authors upon reasonable request.

\header{Code availability} \newline
The code supporting the findings of this study is available from the corresponding authors upon reasonable request.

\header{Acknowledgements} \newline
The authors thank members of the Penn WAVES Lab for their discussions and feedback on the manuscript.
This work is funded in part by NSF 2516558.

\header{Author contributions} \newline
H.L. designed and implemented the system, collected and analyzed the data, and drafted the manuscript.
Z.Z. developed the scene-level field and contributed to data collection, evaluation, and manuscript preparation.
M.Z. contributed to the study conception, supervised the work, and revised the manuscript.
All authors discussed the results and reviewed the final manuscript.

\header{Competing interests} \newline
The authors declare no competing interests.

\header{Additional information} \newline
{\raggedright
Corresponding authors: Haowen Lai (\mbox{\href{mailto:hwlai@seas.upenn.edu}{hwlai@seas.upenn.edu}}) and Mingmin Zhao (\mbox{\href{mailto:mingminz@seas.upenn.edu}{mingminz@seas.upenn.edu}}).
\par}

\addtolength{\bibsep}{-1pt}
\bibliography{bibliography.bib}
\end{document}